\documentclass[a4paper,fleqn]{cas-dc}

\usepackage[authoryear,longnamesfirst]{natbib}
\usepackage{amsmath,amsfonts}
\usepackage{algorithmic}
\usepackage{algorithm}
\usepackage{array}
\usepackage[caption=false,font=normalsize,labelfont=sf,textfont=sf]{subfig}
\usepackage{textcomp}
\usepackage{url}
\usepackage{verbatim}
\PassOptionsToPackage{final}{graphicx}
\usepackage{siunitx}
\usepackage{amssymb}
\usepackage{multirow,booktabs}
\usepackage{threeparttable}
\usepackage{verbatim}
\usepackage{bm}
\usepackage{scalerel}
\usepackage{tikz}
\usepackage{xcolor} 
\usepackage{breakurl}
\usepackage{titlesec}
\usepackage[switch]{lineno}
\definecolor{myblue}{RGB}{33,150,209} 
\hypersetup{colorlinks=true,urlcolor=myblue,linkcolor=myblue,citecolor=myblue} 
 
\titleformat{\section}[block]{\normalsize\bf}{\arabic{section}.}{1em}{}[ ]
\titleformat{\subsection}[block]{\normalsize\it}{\arabic{section}.\arabic{subsection}.}{1em}{}[ ]

\def\tsc#1{\csdef{#1}{\textsc{\lowercase{#1}}\xspace}}
\tsc{WGM}
\tsc{QE}

\renewcommand{\normalsize}{\fontsize{8}{8}\selectfont}
\begin{document}
\let\WriteBookmarks\relax
\def\floatpagepagefraction{1}
\def\textpagefraction{.001}

\shorttitle{Detection of Small Targets in Sea Clutter Based on RepVGG and Continuous Wavelet Transform}    

\shortauthors{Jingchen Ni, Haoru Li, LiLin Xu et al}  

\title [mode = title]{Detection of Small Targets in Sea Clutter Based on RepVGG and Continuous Wavelet Transform}  



%

\author[]{\textcolor{black}{Jingchen} Ni\textcolor{myblue}{$^{a,c}$}}
\author[]{\textcolor{black}{Haoru} Li\textcolor{myblue}{$^{a,c}$}}
\author[]{\textcolor{black}{Lilin} Xu\textcolor{myblue}{$^{a,c}$}}
\author[]{\textcolor{black}{Jing Liang}\corref{cor1}\textcolor{myblue}{$^{b}$}}[orcid=0000-0002-0860-6563]


\ead{liangjing@uestc.edu.cn}



\affiliation[1]{organization={School of Computer Science and Engineering},
            addressline={University of Electronic Science and Technology of China}, 
            city={Chengdu},
            postcode={611731}, 
            state={Sichuan},
            country={China}}
\affiliation[2]{organization={School of Information and Communication Engineering},
            addressline={University of Electronic Science and Technology of China}, 
            city={Chengdu},
            postcode={611731}, 
            state={Sichuan},
            country={China}}
\affiliation[3]{Jingchen Ni, Haoru Li and Lilin Xu are co-first authors.}
\cortext[cor1]{Corresponding author}









\begin{abstract}
    Constructing a high-performance target detector under the background of sea clutter is always necessary and important. In this work, we propose a RepVGGA0-CWT detector, where RepVGG is a residual network that gains a high detection accuracy. Different from traditional residual networks, RepVGG keeps an acceptable calculation speed. Giving consideration to both accuracy and speed, the RepVGGA0 is selected among all the variants of RepVGG. Also, continuous wavelet transform (CWT) is employed to extract the radar echoes' time-frequency feature effectively.  In the tests, other networks (ResNet50, ResNet18 and AlexNet) and feature extraction methods (short-time Fourier transform (STFT), CWT) are combined to build detectors for comparison. The result of different datasets shows that the RepVGGA0-CWT detector performs better than those detectors in terms of low controllable false alarm rate, high training speed, high inference speed and low memory usage. This RepVGGA0-CWT detector is hardware-friendly and can be applied in real-time scenes for its high inference speed in detection.
\end{abstract}



\begin{keywords}
 \sep RepVGG \sep CWT \sep target detector \sep sea clutter
\end{keywords}

\maketitle

\section{Introduction}


Target detection is the main issue of a radar system and raises many detection methods (\cite{fuzzy-logic, semi-supervised}). Detection of sea surface targets is vital in ocean surveillance and sea search-and-rescue field (\cite{vital,field1,field2}). However, establishing a high-performance detector is challenging as the target echo will be covered by sea clutter. Also, the non-Gaussian and non-stationary features of the sea background make small marine target detection even more difficult. Traditional statistical methods try to fit the sea clutter amplitude using non-Gaussian distributions like K-distribution and employing constant false alarm (CFAR) (\cite{CFAR, CFAR2}) or matched filter (MF) (\cite{MF1,MF2}) to detect the target. However, these methods fail to achieve high performance, for it is largely influenced by the sea state. 

Nowadays, researchers try to apply machine learning methods to this topic, which better adapt to real data in clutter modeling. Methods like k-nearest neighbor (KNN) (\cite{KNN}) and support vector machine (SVM) (\cite{mu-SVM}) are applied in this field. KNN is a simple algorithm that classifies the data based on the density of different labeled samples near the tested sample, but it needs to set the K value manually and will probably encounter curse of dimensionality(\cite{curse_of_dimensionality}). SVM maps the data to high-dimensional space and uses hyperplane to divide different classes, but it faces difficulties when dealing with large-scale data. As these methods can't learn the characteristic patterns of the Radar echo thoroughly, deep learning methods are also deployed in this field due to the fast development of chip computing capability. Researchers mainly use sequential features or two-dimensional time-frequency features as the input of the deep learning models. For sequential feature input, Wan used Bi-LSTM and achieves a quite high detection performance (\cite{BiLSTM}). For two-dimensional time-frequency feature input, convolutional neural network (CNN) structures have been the basic backbone of most detectors due to their excellent ability in learning features of two-dimensional data (\cite{CNN}). In 2003, Lopez-Risueno's team deployed short-time Fourier transform (STFT) to extract two-dimensional time-frequency data and employed a simple CNN as the detector (\cite{CNN-2003}). After more complex CNNs are invented, the method of residual convolution neural network combined with two-dimensional Fourier transforms is widely used. For example, Qu used STFT and STDN (\cite{STDN}) while Liu's team used FRFT and ResNet (\cite{Liu}) to construct residual CNN detectors.


Most CNN-based methods tended to use the variants of Fourier transform to extract the time-frequency data inputs. However, most Fourier transforms face the same issue that trigonometric functions with infinite lengths are not adept in capturing sudden variations in sea clutter background, which is significant for target detection. Instead, continuous wavelet transform (CWT) with finite length performs better in capturing these features while working on a multi-scale basis. Also, wavelet transform is widely used in data fusion and data processing in remote sensing.

Although the network with residual blocks like STDN and ResNet can solve the problem of network degradation (\cite{ResNet}), it also reduces the inference speed of the detector in real applications compared to simple VGG-like network structures. To solve this problem, Ding's team proposed RepVGG network by changing residual blocks to equivalent VGG-like structures in the inference stage (\cite{RepVGG}). Also, for multi-resolution wavelet transform, different convolution kernel sizes of RepVGG can obtain multiple sizes of receptive fields, which can better classify time-frequency data built by wavelet transform. 

\begin{figure}[h]
  \centering
  \includegraphics[width = 0.5\textwidth]{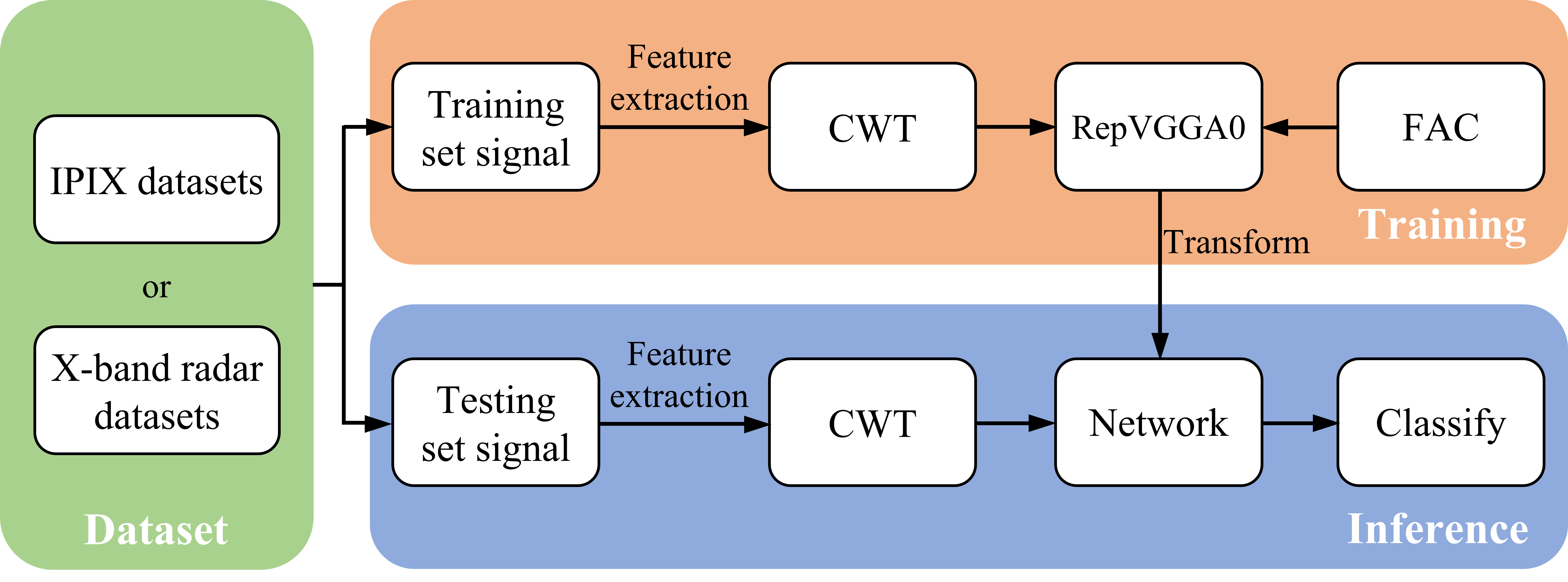}
  \caption{Detection framework of RepVGGA0-CWT detector}
  \label{flow}
  \end{figure}

To build a high-performance CNN-based detector and improve inference speed of the residual networks, we propose a RepVGGA0-CWT detector based on false alarm control (FAC) method (\cite{FAC}).We use CWT method to extract the time-frequency domain feature and use RepVGG to learn the extracted characteristic patterns. It's believed that the information in CWT data that indicates the presence of the target can be mined by the RepVGG network. Considering both accuracy and speed, RepVGGA0 is chosen as the network backbone of the detector. The detector is tested on both IPIX datasets and X-band radar datasets and is proved to be more efficient than traditional CNN-based detectors. The detection framework of the RepVGGA0-CWT detector is shown in \textcolor{myblue}{\textcolor{myblue}{Fig.}} \ref{flow}. 

In summary, the article presents the following contributions:

\begin{enumerate}
  \item We use CWT instead of two-dimensional Fourier transforms to extract informative patterns from primitive data effectively. The RepVGG-like network used in this work performs well in classification tasks while keeping an acceptable inference speed.
  \item The detector combining RepVGG and CWT achieves high detection accuracy with a low false alarm rate, high training speed, high inference speed, and low memory usage. The RepVGG network with multi-size receptive fields is specialized in catching the characteristic differences between sea clutter and targets in multi-resolution CWT data. 
\end{enumerate}

The structure of the paper is as follows. Section \uppercase\expandafter{\romannumeral2} presents the basic mathematical model of this work.  Section \uppercase\expandafter{\romannumeral3} introduces the dataset construction and the feature extraction methods. Section \uppercase\expandafter{\romannumeral4} briefly introduces RepVGG network. Section \uppercase\expandafter{\romannumeral5} illustrates the structure and training configuration of the RepVGGA0-CWT detector. Section \uppercase\expandafter{\romannumeral6} uses both IPIX datasets and recent X-band radar datasets to analyze the detection performance of different detectors (networks, window length and feature extraction methods). Finally, a conclusion of this work is made in Section \uppercase\expandafter{\romannumeral7}.

  The goal of our work is to improve the target detection probability while keeping the false alarm probability low.

\section{Basic Mathematical Model}
  The problem of detecting marine targets can be formulated as a binary hypothesis test, where the null hypothesis $H_0$ represents the absence of a target and the alternative hypothesis $H_1$ represents the presence of a target.
  
  \begin{equation}
    \begin{aligned}
      &H_0:
      \left\{
        \begin{aligned}
          &\bm{x}_k=\bm{c}_k+\bm{n}_k,&k = 1,2,\dots,K\\
          &\bm{x}_k^p=\bm{c}_k^p+\bm{n}_k,&p = 1,2,\dots,P
        \end{aligned}
      \right.
      \\
      &H_1:
      \left\{
        \begin{aligned}
          &\bm{x}_k=\bm{s}_k+\bm{c}_k+\bm{n}_k,&k = 1,2,\dots,K\\
          &\bm{x}_k^p=\bm{c}_k^p+\bm{n}_k,&p = 1,2,\dots,P
        \end{aligned}
      \right.
      \end{aligned}
  \end{equation}

  The radar echo sequence, represented by $\bm{x}$, is composed of several components: the primitive radar echo ($\bm{x}_k$), the target echo component ($\bm{s}_k$), the sea clutter component ($\bm{c}_k$), and the noise component ($\bm{n}_k$). 

  The target detection probability $P_d$ and  false alarm probability $P_{fa}$ are defined below:
  
  \begin{equation}
    \label{definition}
    \begin{aligned}
      &P_d = P(H_1|H_1)\\
    &P_{fa} = P(H_1|H_0)
    \end{aligned}
  \end{equation}

  The goal of our work is to improve the target detection probability while keeping the false alarm probability low.
  \section{Dataset Construction and Feature Extraction}

  \subsection{Dataset Construction}

  In this work, 10 sets of IPIX datasets are used, which were measured in Dartmouth in 1993 under different sea clutter environments (\cite{IPIX}). The basic information of 10 datasets is shown in \textcolor{myblue}{Table} \ref{datasets}. The target is located at the primary target cell. The secondary target cells are the cells close to the primary target cell that are impacted by the target's energy.

  Each dataset consists of four different polarization modes: HH, VV, HV, and VH. Each polarization mode includes 14 distance units. These units are composed of one primary target unit, two to three secondary target units and the remaining units are sea clutter units. Each distance unit contains $2^{17}$ time series, which are sampled at a frequency of 1000 Hz, and has a duration of approximately 131 seconds.

  To split the data, the following formula was used:
  
  \begin{equation}
    \label{form3}
  x_i = x[M\cdot(i-1)+1:M\cdot(i-1)+N]\qquad i = 1,2,...
  \end{equation}
  where $M$ represents the interval between the observation windows and $N$ represents the length of the observation windows.

  \begin{figure}[h]
    \centering
    \includegraphics[width=3in]{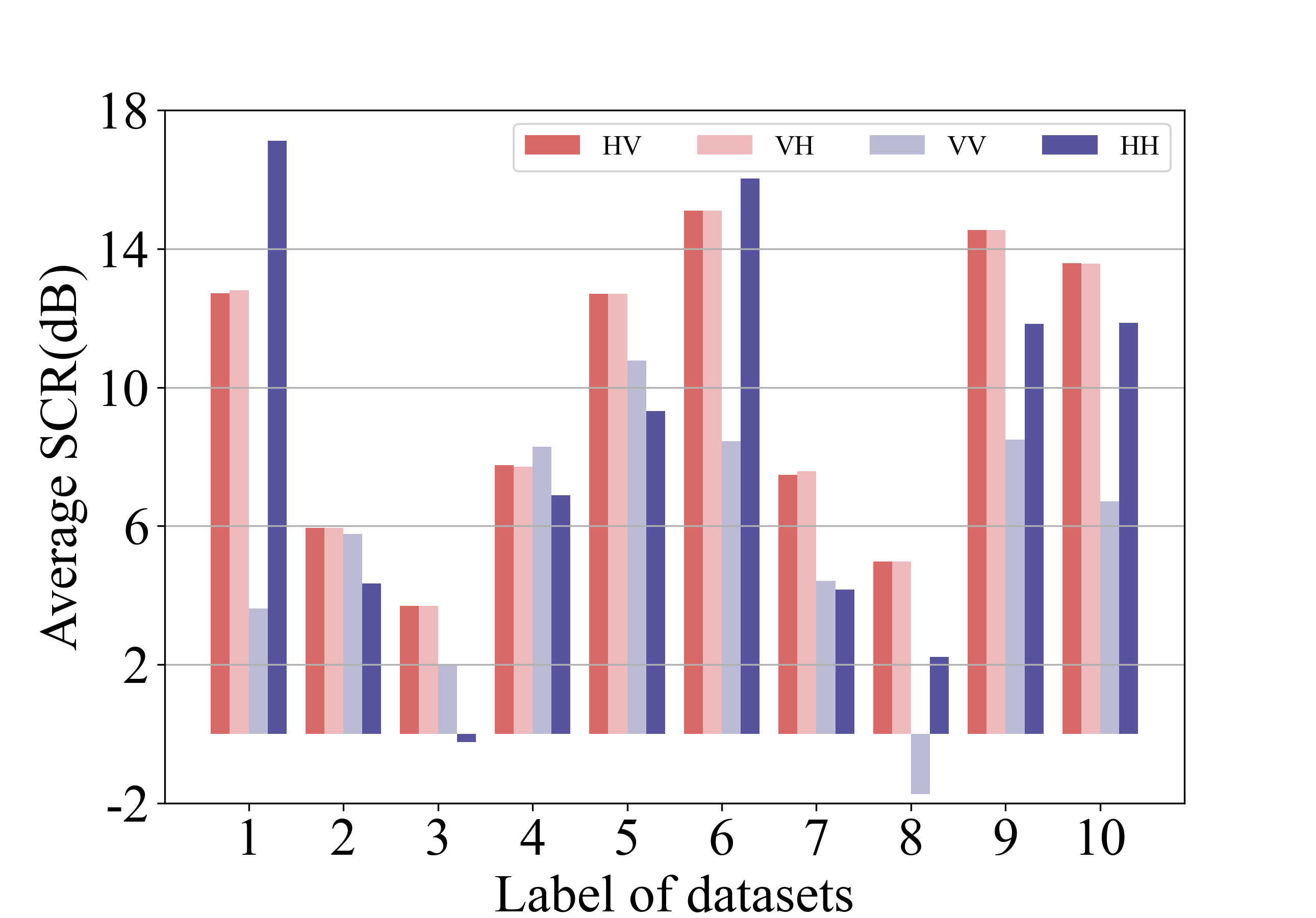}
    \caption{Average SCR of each dataset at primary target cell}
    \label{scr}
  \end{figure} 

The average signal-to-clutter ratio (ASCR) (\cite{BiLSTM}) of a target cell can be computed as

\begin{equation}
\text{ASCR}=10\log_{10}(\frac{2^{-17}\sum_{i=1}^{2^{17}}|x(i)|^2-\overline{P_c}}{\overline{P_c}})
\end{equation}
where $2^{17}$ is the length of one echo cell in the IPIX dataset, $x(i)$ represents the echo data of the primary target cell, and $\overline{P_c}$ represents the average power of pure clutter cell. The average SCRs of ten datasets at primary target cells is shown in \textcolor{myblue}{\textcolor{myblue}{Fig.}} \ref{scr}.

  The radar data at different distances are divided into labeled samples using the \textcolor{myblue}{Eq.} \ref{form3}. Target samples are labeled as 1 and clutter samples are labeled as 0. The samples are then randomly shuffled and divided into training and test datasets, with $70\%$ and $30\%$ of the samples being used for each dataset respectively.

  To ensure the high detection speed, the number of points in the frequency domain of STFT is set to $128$ and the length of the window function is also set to $128$. For CWT, a subset is selected from the CWT result evenly to obtain a subset whose longest dimension length does not exceed $128$.

  In deep learning using convolutional neural networks, concatenation is a commonly used method for data fusion. In \textcolor{myblue}{Table} \ref{polarizationAVG}, the experiments in HV and VH polarizations performs better than other polarizations. Therefore, we employ concatenation to combine two-dimensional data from the HV and VH polarizations to obtain HV,VH data. The concatenation process is shown below: 
  \begin{figure}[h]
    \centering
    \includegraphics[width=3in]{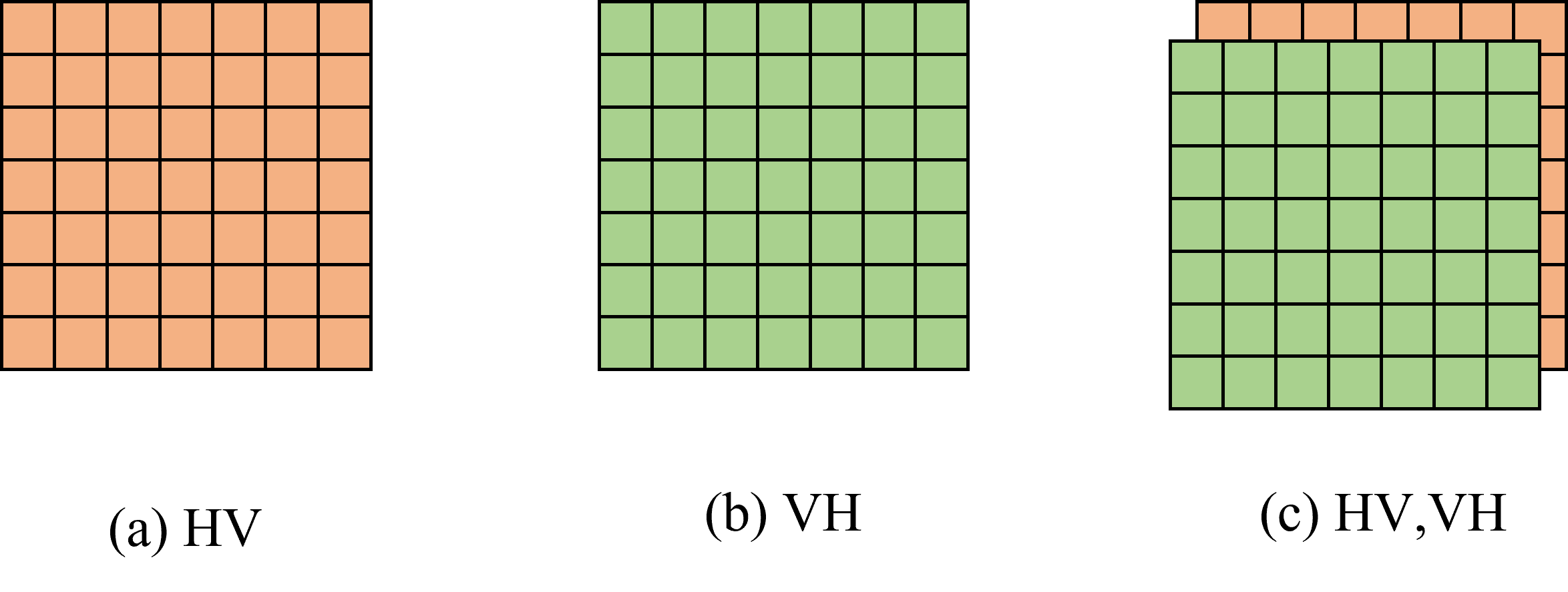}
    \caption{Concatenation of HV and VH data}
    \label{concatenation}
    \end{figure}  
  \begin{table*}
    \centering
    \renewcommand\arraystretch{1.5}
    \caption{primitive data selected from 1993 IPIX dataset}
    \label{datasets}
    \begin{threeparttable}
    \begin{tabular}{c c c c c c c}
      \toprule
    Label & File Name                    & Primary target & Secondary target & Sea state level & SWH(m) & WS(km/h)  \\
    \midrule
    1                         & 19931107\_135603\_starea     & 9                                  & 8,9,10,11        & 4                                   & 2.2                        & 9                             \\
    2                         & 19931108\_220902\_starea     & 7                                  & 6,7,8            & 3                                   & 1.1                        & 9                             \\
    3                         & 19931109\_191449\_starea     & 7                                  & 6,7,8            & 2                                   & 0.9                        & 19                            \\
    4                         & 19931109\_202217\_starea     & 7                                  & 6,7,8,9          & 2                                   & 0.9                        & 19                            \\
    5                         & 19931110\_001635\_starea     & 7                                  & 5,6,7,8          & 2                                   & 1                          & 9                             \\
    6                         & 19931111\_163625\_starea     & 8                                  & 7,8,9,10         & 2                                   & 0.7                        & 20                            \\
    7                         & 19931118\_023604\_stareC0000 & 8                                  & 7,8,9,10         & 3                                   & 1.6                        & 10                            \\
    8                         & 19931118\_162155\_stareC0000 & 7                                  & 6,7,8,9          & 2                                   & 0.9                        & 33                            \\
    9                         & 19931118\_162658\_stareC0000 & 7                                  & 6,7,8,9          & 2                                   & 0.9                        & 33                            \\
    10                        & 19931118\_174259\_stareC0000 & 7                                  & 6,7,8,9          & 2                                   & 0.9                        & 28                           \\ \bottomrule
    \end{tabular}
    \begin{tablenotes}
    \footnotesize
    \item[1] SWH denotes significant wave height.

    \item[2] WS denotes wind speed.
      
    \item[3] The sea state level adopts the Douglas Sea Division Standard \cite{IPIX}.
    \end{tablenotes}
  \end{threeparttable}
    \end{table*}

  \subsection{STFT}
When dealing with a signal that has a changing frequency, it is appropriate to use the short-time Fourier transform (STFT) to extract its features as it incorporates both time and frequency information. The STFT is defined mathematically as:

\begin{equation}
G(t,f)=\int_R x(\tau)w(\tau-t)e^{-j2\pi f\tau}\ \text{d}\tau
\end{equation}
Where $x(\tau)$ is the original signal and $w(\tau)$ is the window function. $G(t,f)$ represents the frequency spectrum of the signal $x(\tau)$ at time t.

For discrete signals, the STFT is defined as:

\begin{equation}
G(m,\omega)=\sum_{k=ms}^{ms+\Omega-1}x(k)w(k-ms)e^{-j\frac{2\pi k\omega}{\Omega}}
\end{equation}

\begin{figure*}[!t]
\centering
\subfloat[]{\includegraphics[width=3.2in]{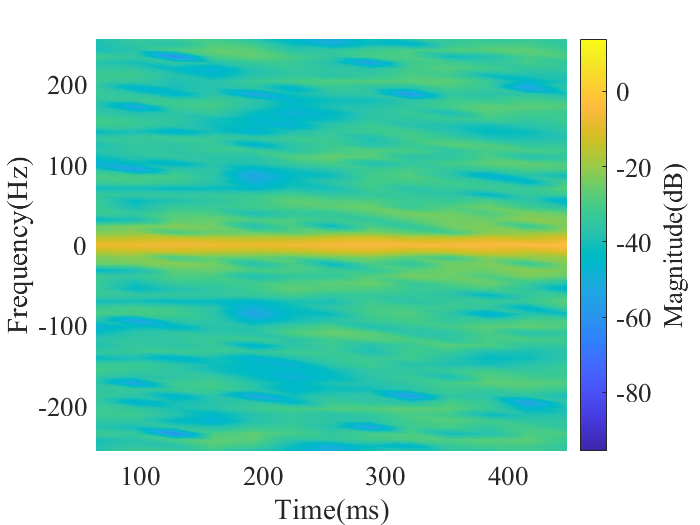}%
\label{STFTc}}
\hfil
\subfloat[]{\includegraphics[width=3.2in]{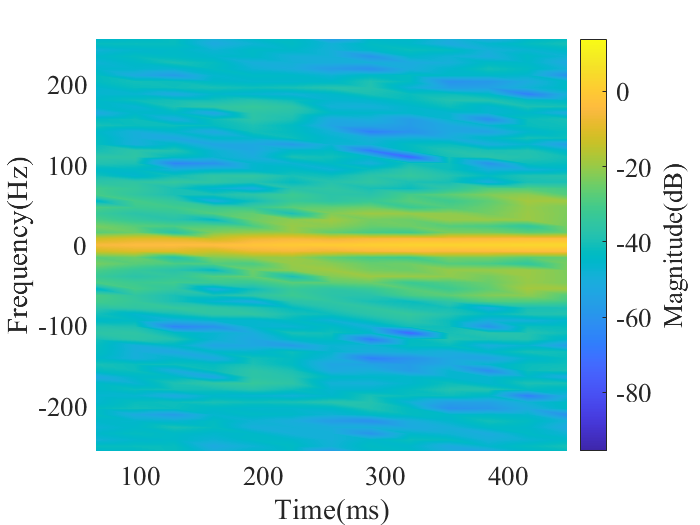}%
\label{STFTt}}
\caption{STFT of (a) sea clutter sample and (b) target sample in No.1 IPIX dataset (VH polarization).}
\label{STFT}
\end{figure*}

where $m$ is the the number of windows, and $\omega$ is the frequency parameter.

The STFT can be illustrated visually, as seen in \textcolor{myblue}{\textcolor{myblue}{Fig.}} \ref{STFT}, which shows two STFT time-frequency images. \textcolor{myblue}{\textcolor{myblue}{Fig.}} \ref{STFTc} is STFT of a sea clutter sample, and \textcolor{myblue}{\textcolor{myblue}{Fig.}} \ref{STFTt} is STFT of a target sample.

\subsection{CWT}

The Continuous Wavelet Transform (CWT) is generally considered to be more effective than the short-time Fourier transform (STFT) in processing non-stationary time series. This is because the STFT is limited by its fixed window function size (\cite{wavelet1}), while the wavelet function used in the CWT is dilated and translated to produce a family of functions with varying scales, making CWT a multi-resolution method(\cite{wavelet1_1}).

The formula for the discrete form of the CWT is as follows:

\begin{equation}
W_{\beta,s}=\int x(\tau)\frac{1}{\sqrt{s}}\Psi(\frac{\tau-\beta}{s})d\tau
\end{equation}
where $x(\tau)$ is the input signal, $s$ is the scale factor, and $\tau$ represents the time shift of the signal. In this work, the Morse wavelet is chosen as the wavelet function $\Psi$, which is defined in the frequency domain as:

\begin{equation}
\Psi_{P,\gamma}(\omega)=U(\omega)a_{P,\gamma}\omega^{\frac{P^2}{\gamma}}e^{-\omega^\gamma}
\end{equation}
where $U(\omega)$ is the unit step function, $a(P,\gamma)$ is a normalizing constant, $P^2$ is the time-bandwidth product, and $\gamma$ characterizes the symmetry of the wavelet. To achieve a symmetric, nearly Gaussian, and time-frequency concentrated member in the Morse wavelet family, the values of $\gamma=3$ and $P^2=60$ are selected (\cite{wavelet2}).

In the case of a signal with a finite length, the convolution operation near the signal borders causes the wavelet window to extend into a region where there is no available data. As a result, the transform values close to the signal borders are affected by non-existent data, leading to abnormal values in the CWT coefficients (\cite{edgeeffect}). The phenomena above is called edge effect.

\begin{figure*}[!t]
\centering
\subfloat[]{\includegraphics[width=3.25in]{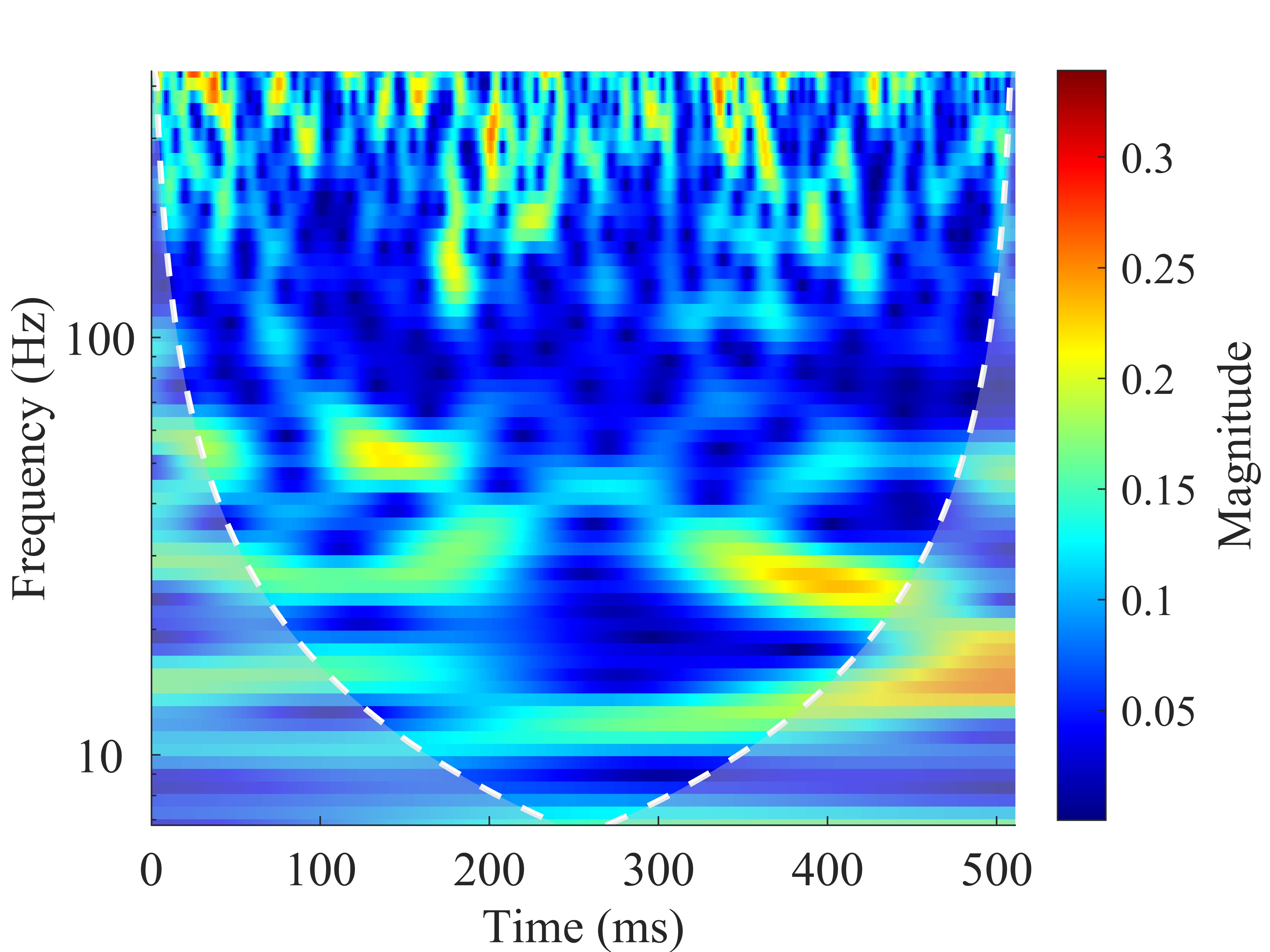}%
\label{CWTc}}
\hfil
\subfloat[]{\includegraphics[width=3.25in]{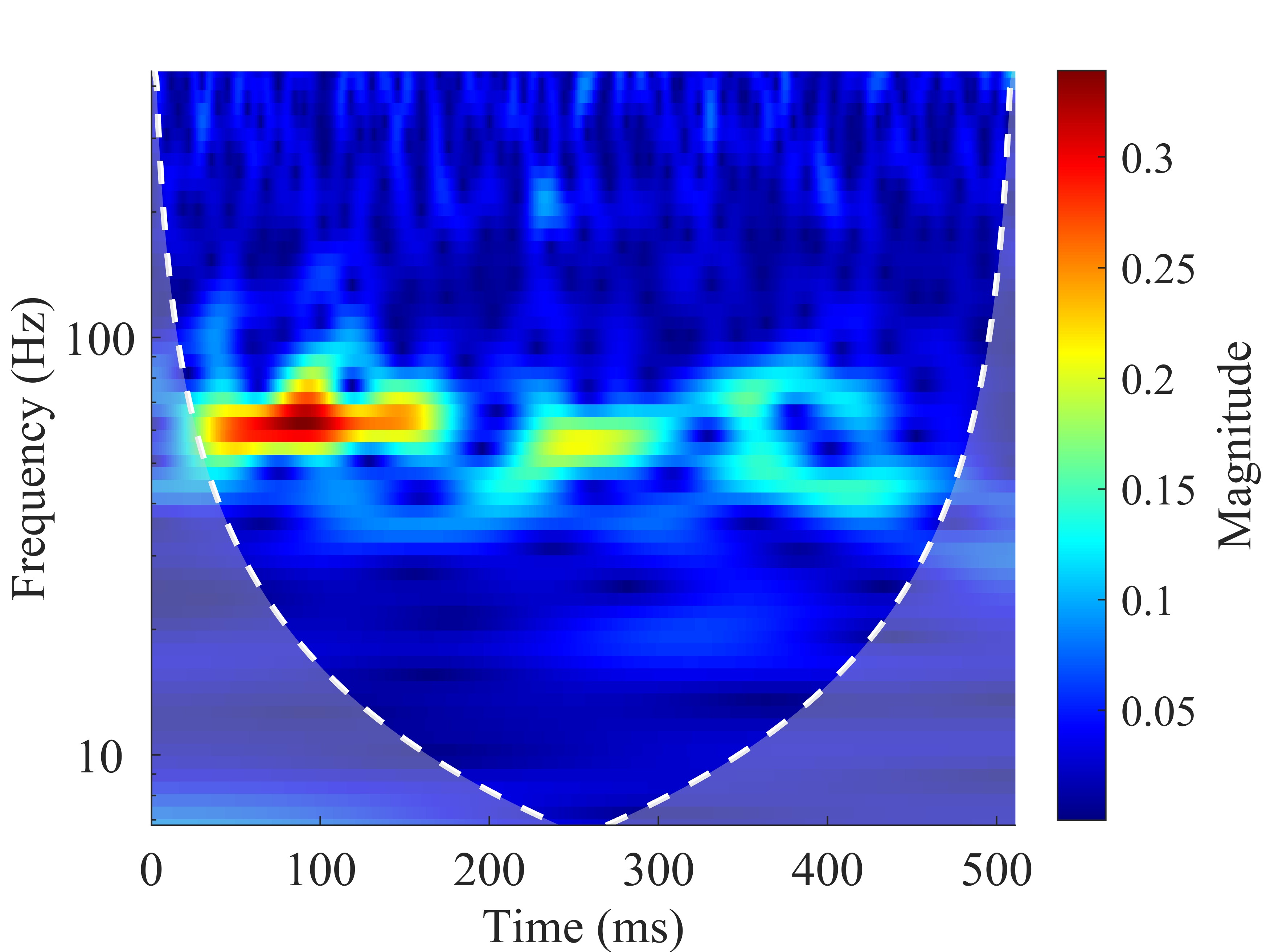}%
\label{CWTt}}
\caption{CWT of (a) sea clutter sample and (b) target sample in No.1 IPIX dataset (VH polarization).}
\label{CWT}
\end{figure*}

Two examples of CWT scalograms are shown in \textcolor{myblue}{\textcolor{myblue}{Fig.}} \ref{CWT}. The gray regions outside the dashed white line represent areas where edge effects are significant. The gray region of high frequency domain is small, and the gray region of low frequency domain is large. This is because wavelet windows in high frequency domain are narrow as wavelet functions are compressed and wavelet windows in low frequency domain are wide because wavelet functions are stretched. The CWT provides a highly recognizable feature that can be learned by neural networks.

\section{RepVGG Network}

In order to improve the training and inference speed of the convolutional networks which have 
residual structures, RepVGG was proposed by Ding's team in 2021. The training network of RepVGG
also applies the idea of ResNet as it links the frontal layer with the lateral layer. When it comes to the inference network, the network has a VGG-like inference body and is reconstructed using a  re-parameterization technique making it use less memory and calculate faster. In \cite{RepVGG}, RepVGG was proved to be faster and  more accurate than ResNet and EfficientNet.

\begin{figure}[h]
  \centering
  \includegraphics[width=3in]{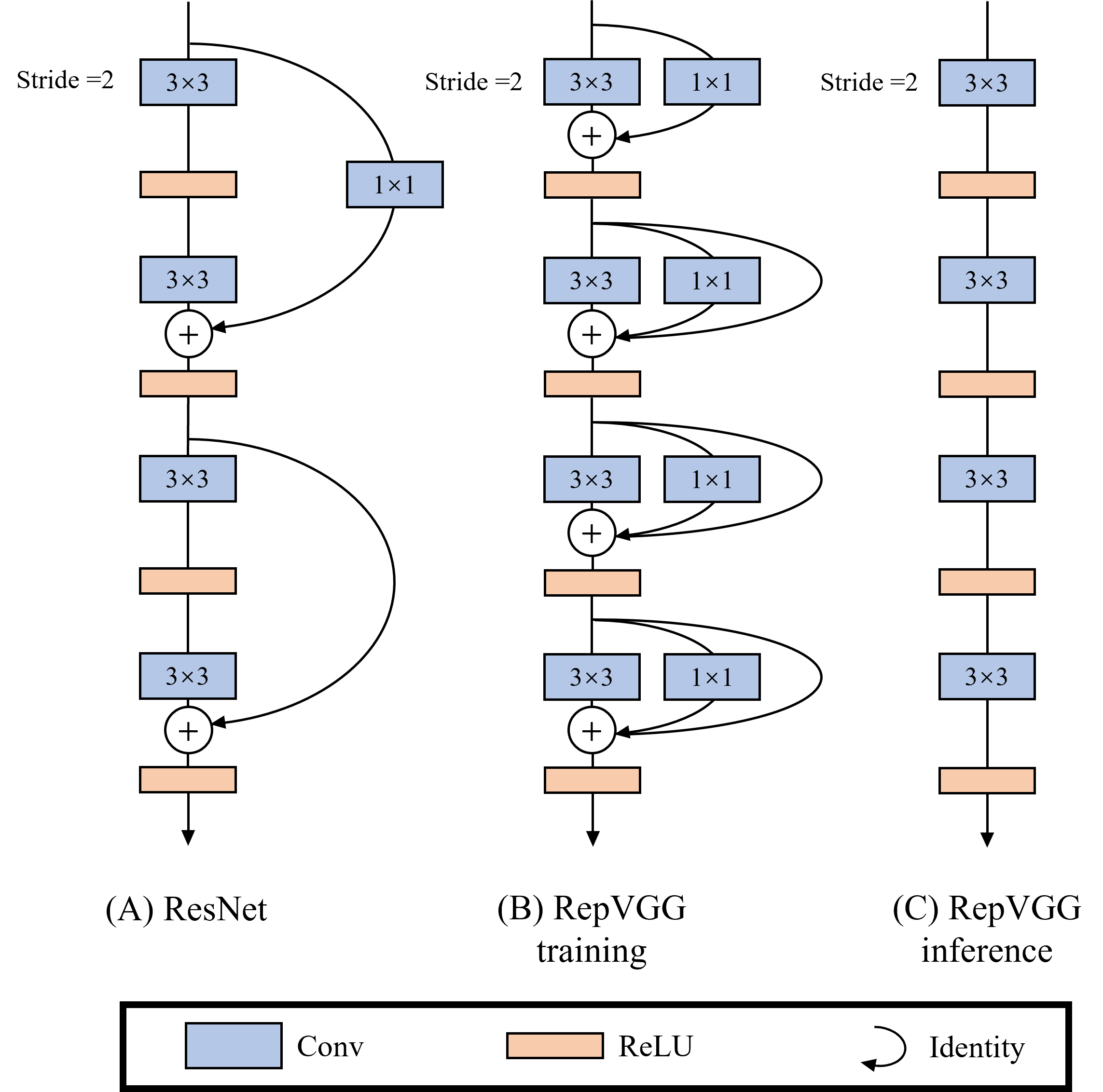}
  \caption{Network structure of RepVGG and ResNet}
  \label{RepVGG}
  \end{figure}

The structure of ResNet and RepVGG training and inference net is shown in \textcolor{myblue}{\textcolor{myblue}{Fig.}} \ref{RepVGG}.

For the RepVGG training net, compared to the ResNet, residual blocks in RepVGG network do not cross layer. Also,
the whole network contains two residual structures. The first layer only contains convolution $1\times1$ residual branch, while the
lateral layers contain not only $1\times1$ residual branch but also identity residual branch. Applying more residual
branches makes the network have multiple gradient flow paths. In the initial stage of the network, a simple residual structure 
is used.
With the deepening of the network, a complex residual structure is used. This not only obtain a more robust 
feature representation in the deep layer of the network, but also deal with the gradient disappearance problem better
in the deep layer of the network.

For residual blocks, the final results can be obtained only after all residual branches have calculated the corresponding 
results. The intermediate results of these residual branches will be stored in the memory of the device, which will have 
a great demand on the memory of the inference device. The round-trip memory operation will reduce the inference speed of 
the entire network. In the RepVGG inference stage, the network is first converted into a single branch $3\times3$ convolution structure offline, which improves the utilization rate of the device memory and enhances the inference 
speed of the network.

\begin{figure}[h]
  \centering
  \includegraphics[width=3in]{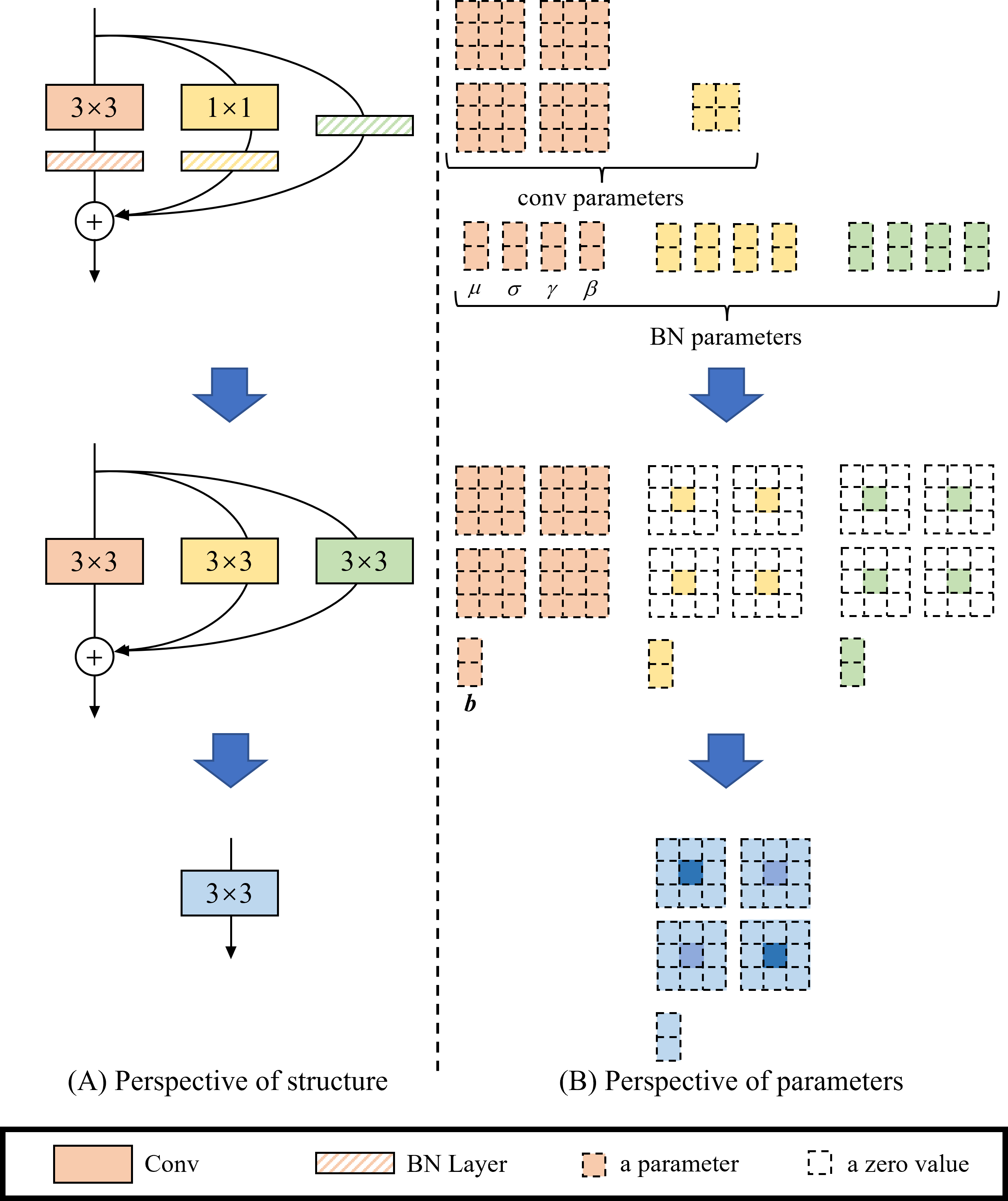}
  \caption{Change of the RepVGG network in inference stage}
  \label{change}
  \end{figure}

To transfer residual blocks to simple $3\times3$ convolution blocks, re-parameterization methods below should be taken, which is the reason why the
network is called RepVGG. \textcolor{myblue}{\textcolor{myblue}{Fig.}} \ref{change} shows the network change of two channels input images from training to inference stage.

The first step is to combine the convolution layer and batch normalization (BN) layer in the residual block.

The operation of the convolutional layer is:

\begin{equation}
Conv(x)=Wx+b
\end{equation}

Here bias isn't used, so the operation becomes:

\begin{equation}
Conv(x)=Wx
\end{equation}

The operation of BN layer (\cite{BN}) is:

\begin{equation}
BN(x)=\gamma\frac{x-Mean}{\sqrt{Var}}+\beta
\end{equation}
where $\gamma$ is a learnable parameter that scales the normalized value, and $\beta$ is a learnable parameter that shifts the normalized value. Both of these parameters can be learned by the network during training.

Combine these two operations:

\begin{equation}
\begin{aligned}
BN(Conv(x))=&\gamma\frac{Wx-Mean}{\sqrt{Var}}+\beta\\
&=\gamma\frac{Wx}{\sqrt{Var}}+\beta-\gamma\frac{Mean}{\sqrt{Var}}
\end{aligned}
\end{equation}

The second step is to change identity layer to $3\times3$ convolution shown in \textcolor{myblue}{\textcolor{myblue}{Fig.}} \ref{change}. In the corresponding $3\times3$ convolution, each kernel has one 1 weight at the center of a different channel.

After the transforms above, these $3\times3$ convolutions are added up to form the equivalent $3\times3$ convolution block.

\begin{figure}[h]
  \centering
  \includegraphics[width=3in]{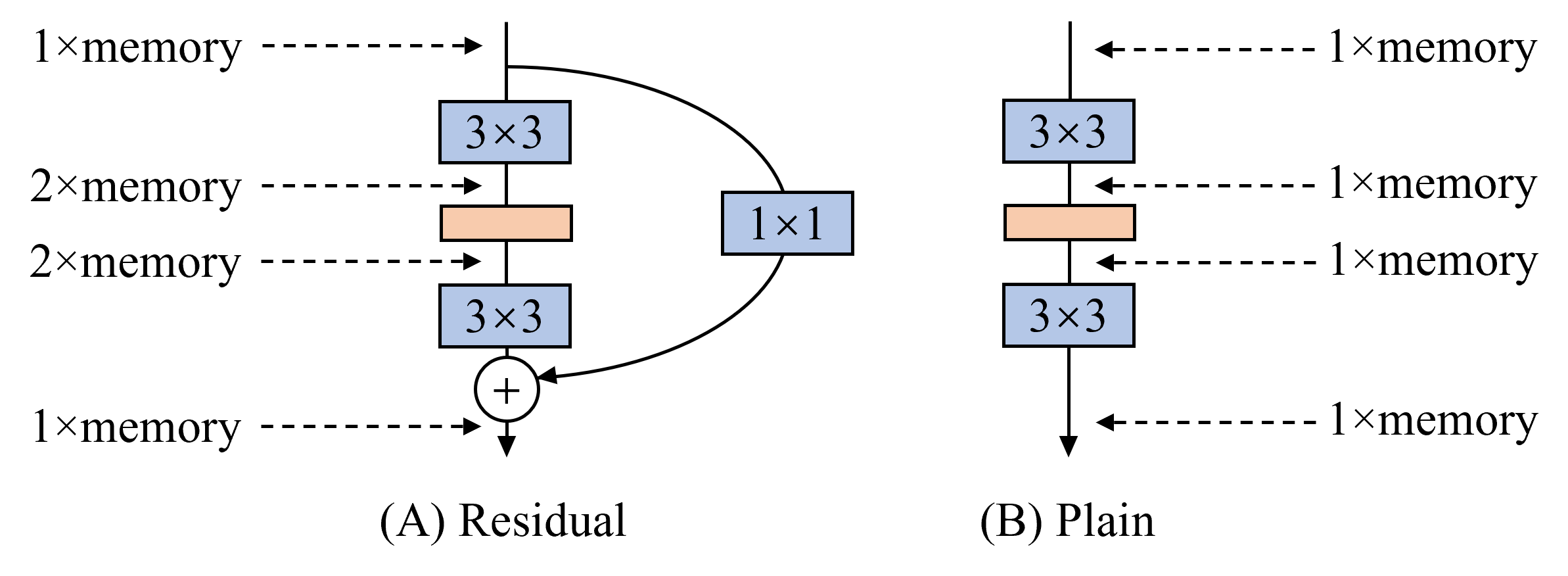}
  \caption{Reduction of memory usage}
  \label{memory}
  \end{figure}

For the reduction of memory usage shown in \textcolor{myblue}{\textcolor{myblue}{Fig.}} \ref{memory}, memory usage is lower than the residual structured network in the inference stage.

In conclusion, RepVGG has these advantages below according to \cite{RepVGG}:
\begin{itemize}
  \item RepVGG is a VGG-like network which means it has no branch and takes the output of its frontal layer as its input.
  \item The body of the network only applies $3\times3$ convolution and ReLU activation function. Using a smaller convolution kernel can reduce the number of parameters while keeping the accuracy of the network. The limited receptive field of two stacked
  $3\times3$ convolution layers is $5\times5$ while the receptive field of three stacked $3\times3$ convolution layers are $7\times7$. So the stacking of small-sized convolutional layers can replace large-sized convolution layers and maintain the same receptive field. Also, $3\times3$ convolution is hardware friendly (\cite{HA}).

\item RepVGG owns receptive fields with different sizes due to its different convolution kernels. This characteristic helps RepVGG excel in processing multi-resolution data such as CWT data.  

  \item The network doesn't contain structures that require heavy designs such as automatic search (\cite{autosearch}), manual refinement or compound scaling (\cite{comscale}).
\end{itemize}

\section{RepVGGA0-CWT Detector}
In this work, we choose RepVGGA0 as the detector backbone for it has a proper inference speed and owns satisfactory performance. The working framework of RepVGGA0-CWT detector is shown in \textcolor{myblue}{\textcolor{myblue}{Fig.}} \ref{flow} . The training set signal is first processed by CWT to obtain time-frequency data. After that, the RepVGGA0 network uses the training time-frequency data to adjust the network parameters and learn the internal pattern of the time-frequency data. Then, FAC method determines the output threshold of the controlled $P_{fa}$ according to the output of network on training data. Finally, the testing signal is processed and tested by the trained model to validate the detection performance of the detector on the test set.
\subsection{Network Training}

The RepVGGA0 network is constructed using the PyTorch framework, with the following parameter settings: a batch size of 128, a maximum number of 25 epochs, 75 iterations per epoch, the Adam optimizer, and an initial learning rate of 0.001. The network parameters are optimized through iterations. To improve the convergence of the network, an exponential learning rate decay is applied to automatically decrease the learning rate in each epoch. The drop rate is set to 0.9.

The loss function used in this work is weighted cross-entropy loss function, considering the imbalance of the amount between target samples and clutter samples (\cite{loss}). The formula of the loss function applied in this work is

\begin{equation}
Loss=-\frac{1}{n}\sum_{i=1}^n wt_i\log(P_i)
\end{equation}
where $n$ represents the total number of classes. $t_i$ is a one-hot encoded vector indicating the label of each sample. $P_i$ is the output of the softmax layer, which represents the probability of each sample belonging to a given class. Most importantly, $w$ is a weight vector whose value is adjustable for different classes. In this work, to solve class imbalance problem, $w$ for target samples is set to 20 and $w$ for clutter samples is set to 1.

\subsection{False Alarm Rate Control}
A previous study proposed a method for controlling the false alarm rate using the output of the softmax layer (\cite{FAC}). The method involves sorting the first item in the output array of the softmax layer, which represents the probability of no target, in ascending order, and then selecting a detection threshold $\eta$ based on the desired false alarm rate.

\begin{equation}
\begin{aligned}
&\eta=O_1(i)\\
&i=Pfa_{desired}\cdot N_{clutter}
\end{aligned}
\end{equation}
where $O_1$ is the first item of the output array of the clutter samples sorted in ascending order and $N_{clutter}$ represents the number of clutter samples. In this work, the threshold $\eta$ is selected based on the results of the test dataset for binary class classification.

\begin{figure*}[!t]
  \centering
  \subfloat[]{\includegraphics[width=0.5\textwidth]{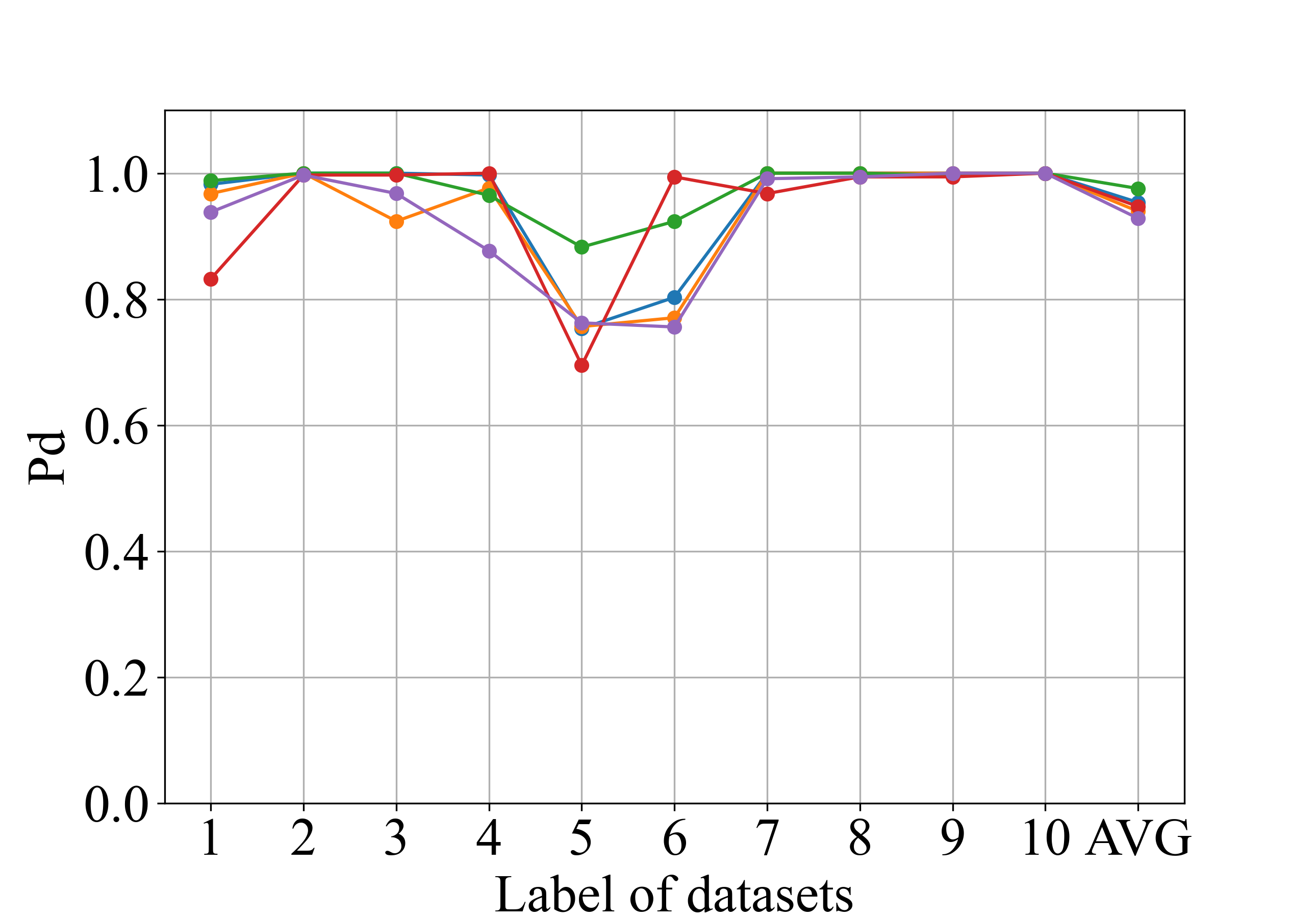}%
  \label{Rep_CWT}}
  \hfil
  \subfloat[]{\includegraphics[width=0.5\textwidth]{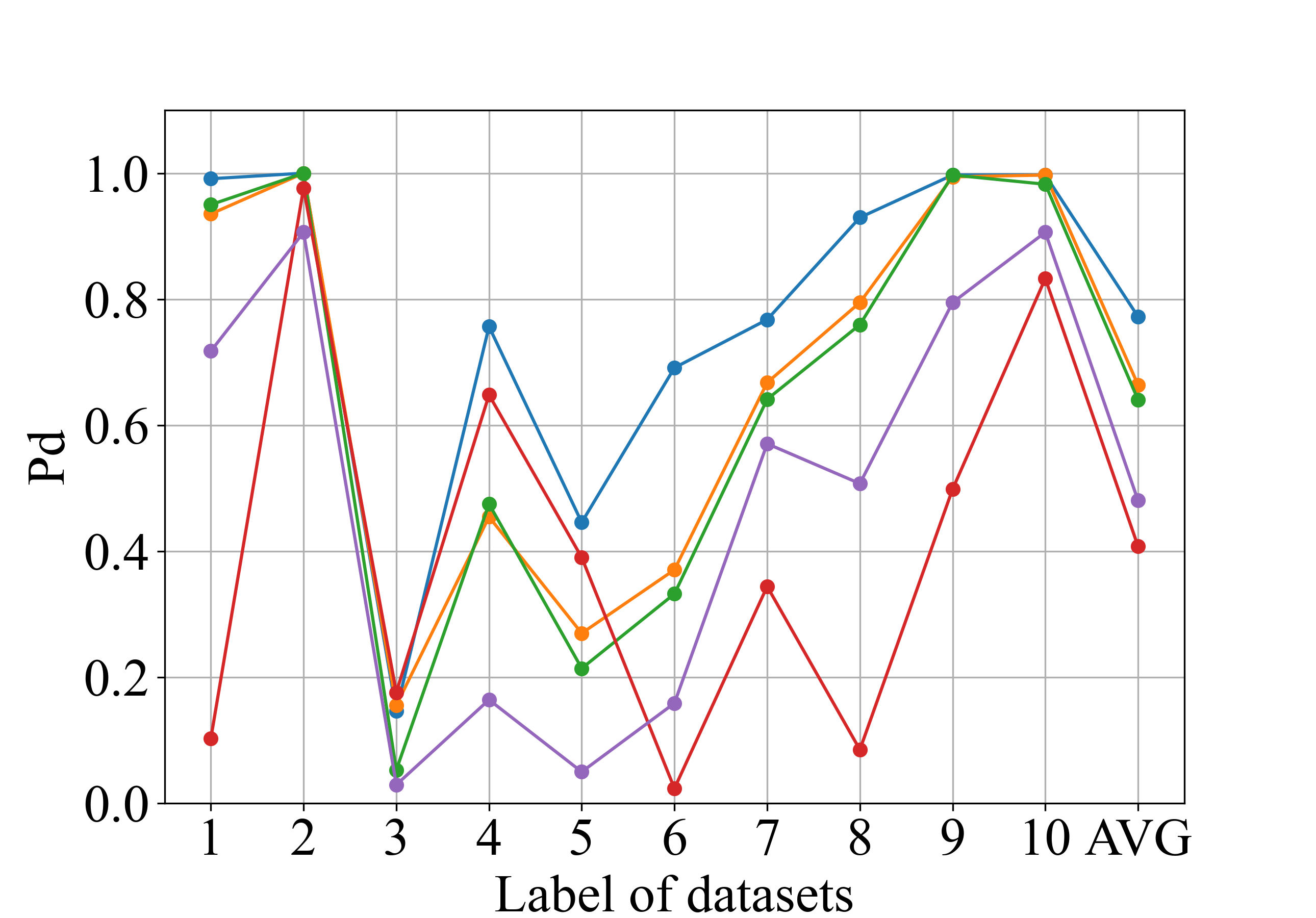}%
  \label{Rep_STFT}}
  \hfil
  \subfloat[]{\includegraphics[width=0.5\textwidth]{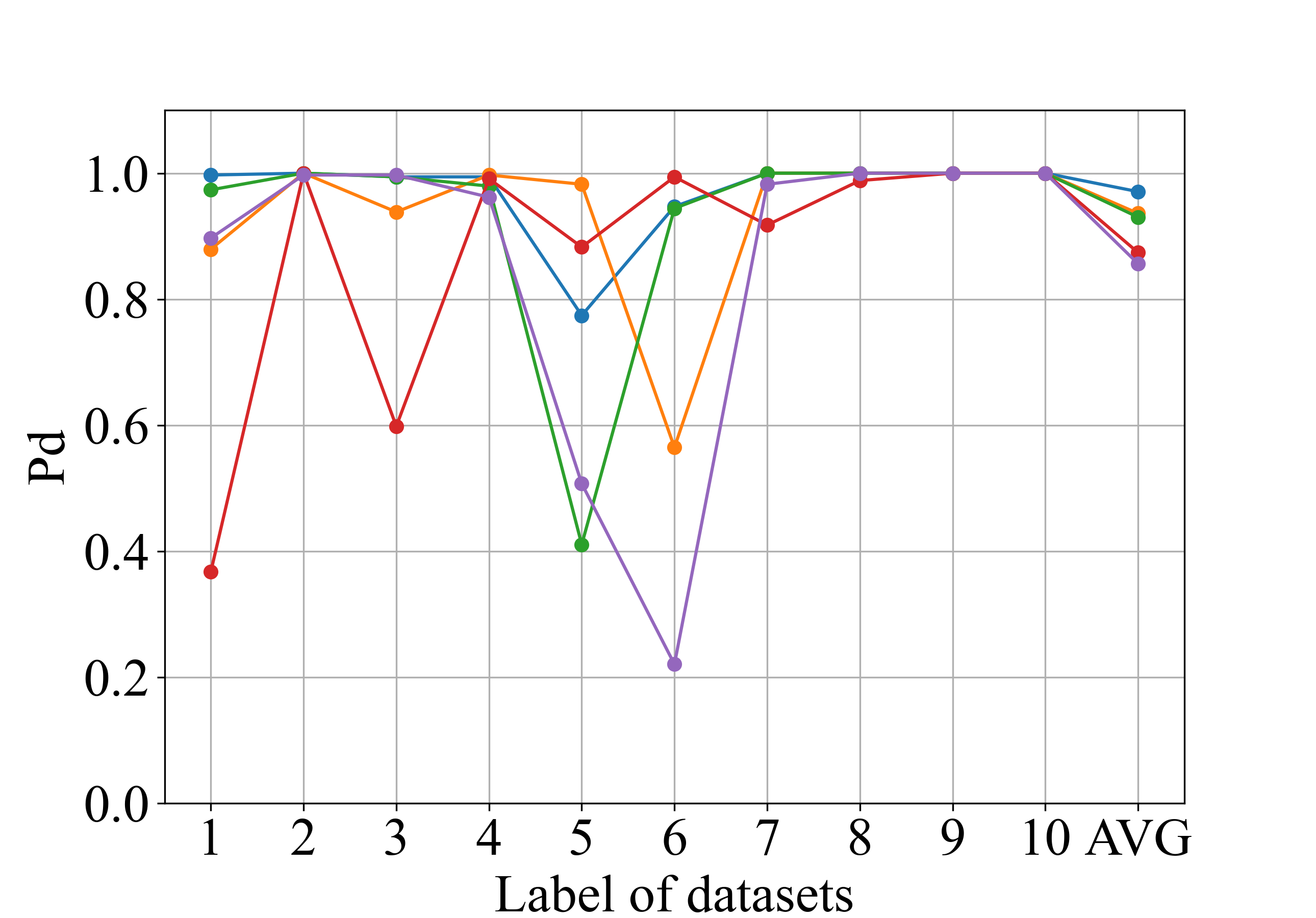}%
  \label{Res_CWT}}
  \hfil
  \subfloat[]{\includegraphics[width=0.5\textwidth]{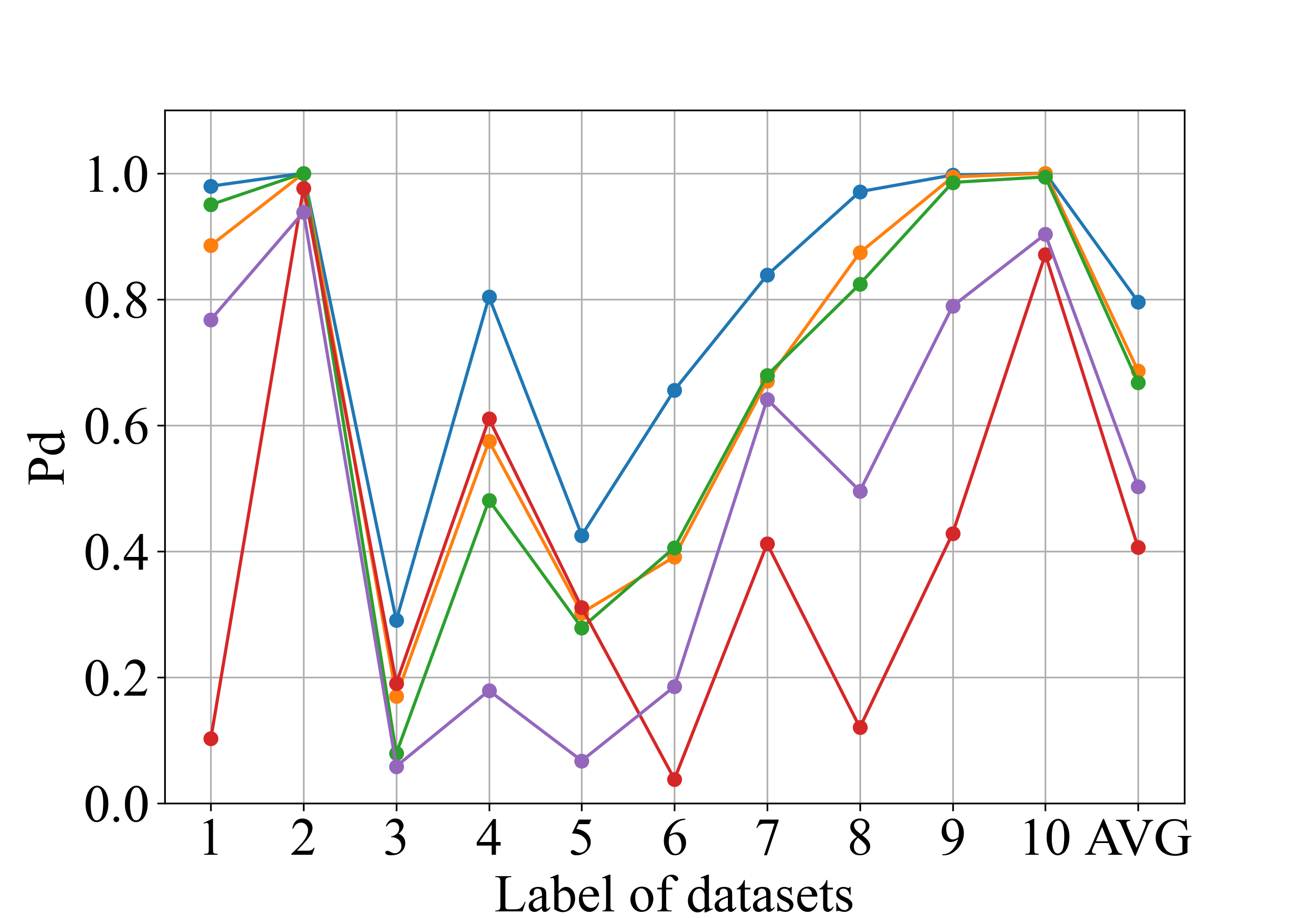}%
  \label{Res_STFT}}
  \hfil
  \includegraphics[width=2in]{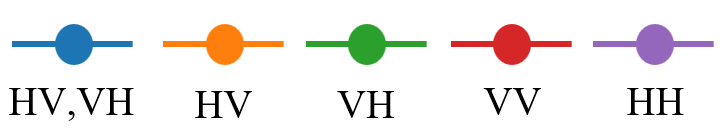}
  \caption{$P_d$ of 4 kinds of detetors of each IPIX dataset when target $P_{fa}$ is set to 0.001 (a)RepVGGA0-CWT. (b)RepVGGA0-STFT. (c)ResNet50-CWT (d)ResNet50-STFT}
  \label{1993_datasets}
  \end{figure*}

\begin{figure*}[!t]
  \centering
  \subfloat[]{\includegraphics[width=0.5\textwidth]{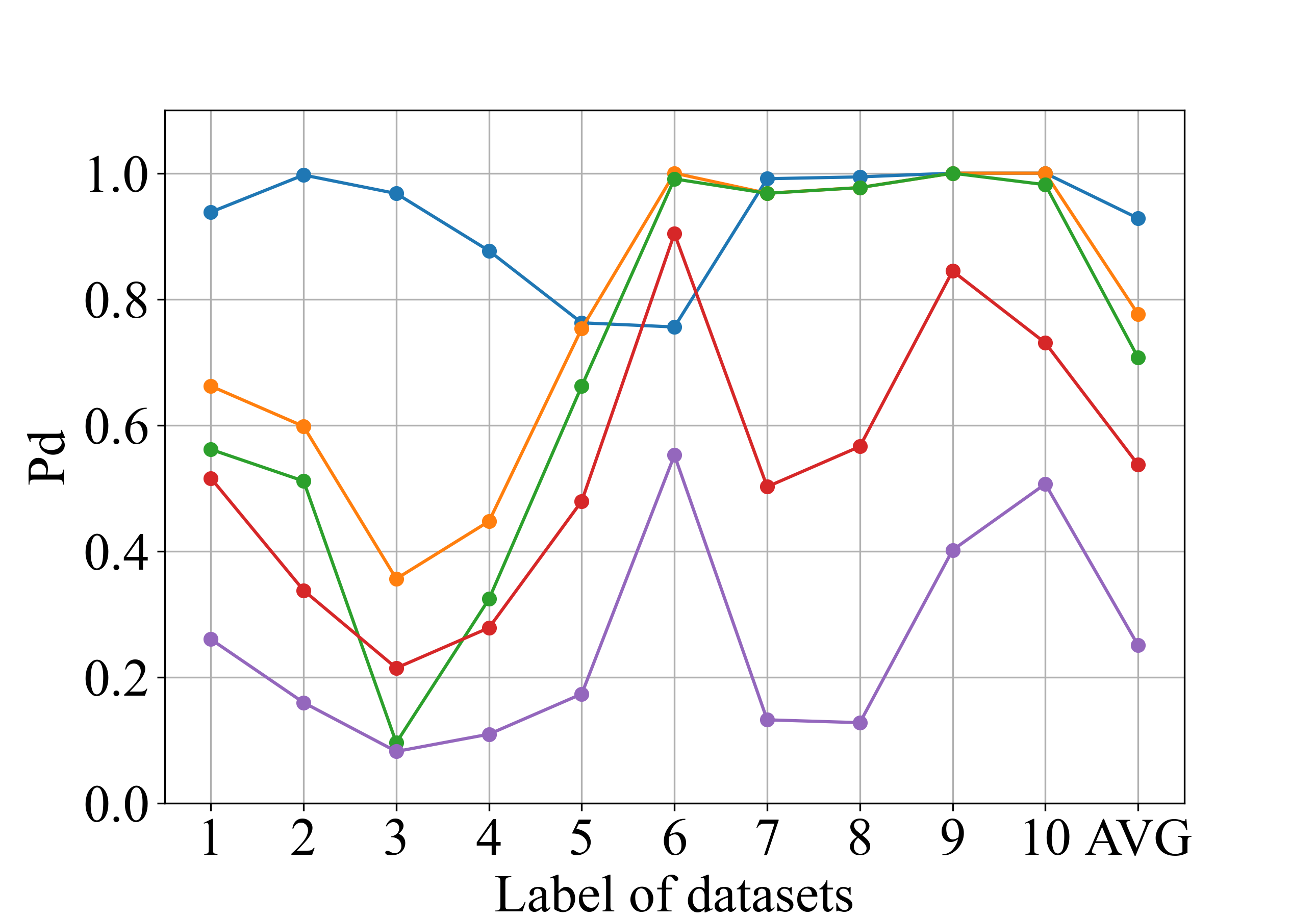}%
  \label{Rep_CWT}}
  \hfil
  \subfloat[]{\includegraphics[width=0.5\textwidth]{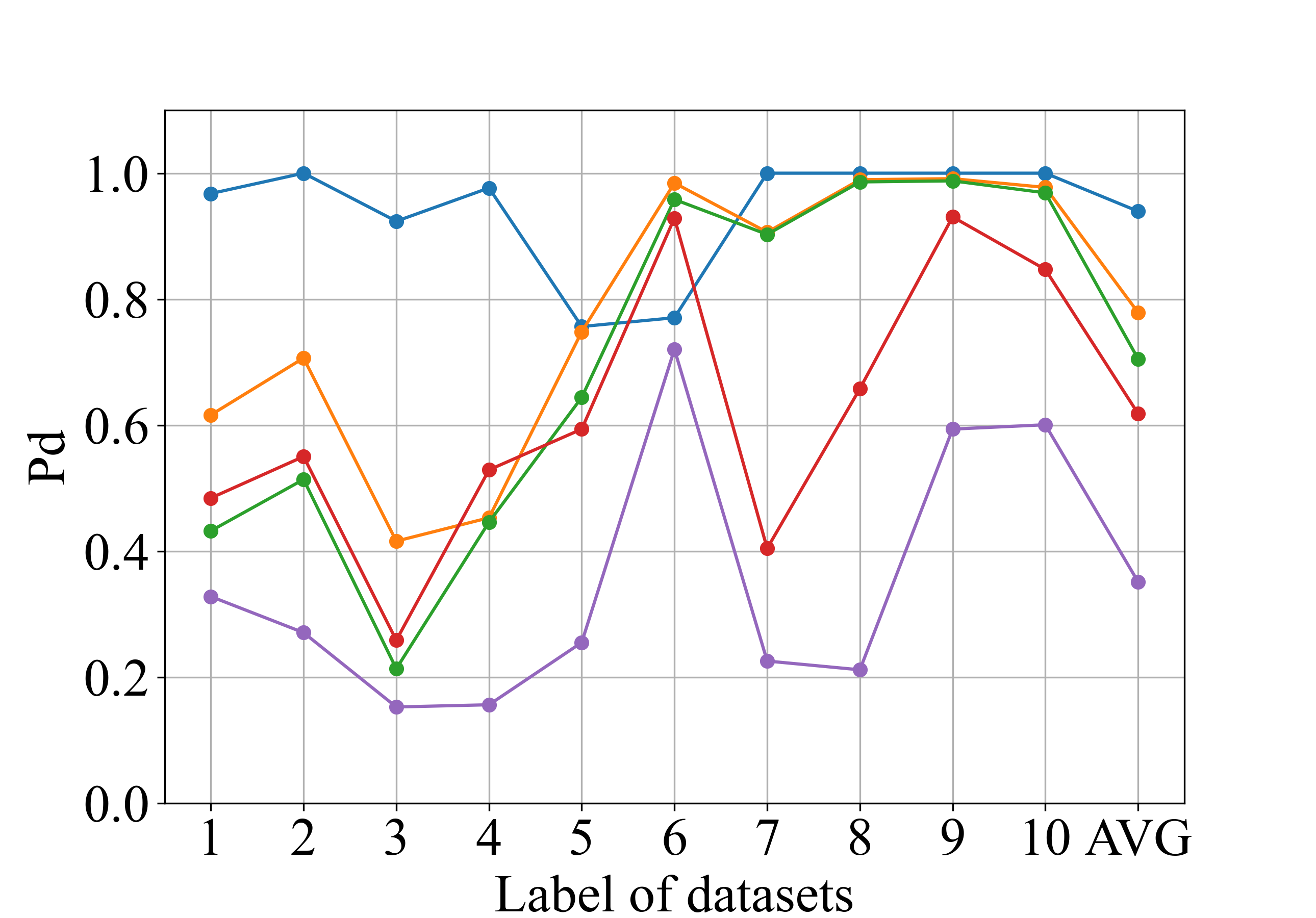}%
  \label{Rep_STFT}}
  \hfil
  \subfloat[]{\includegraphics[width=0.5\textwidth]{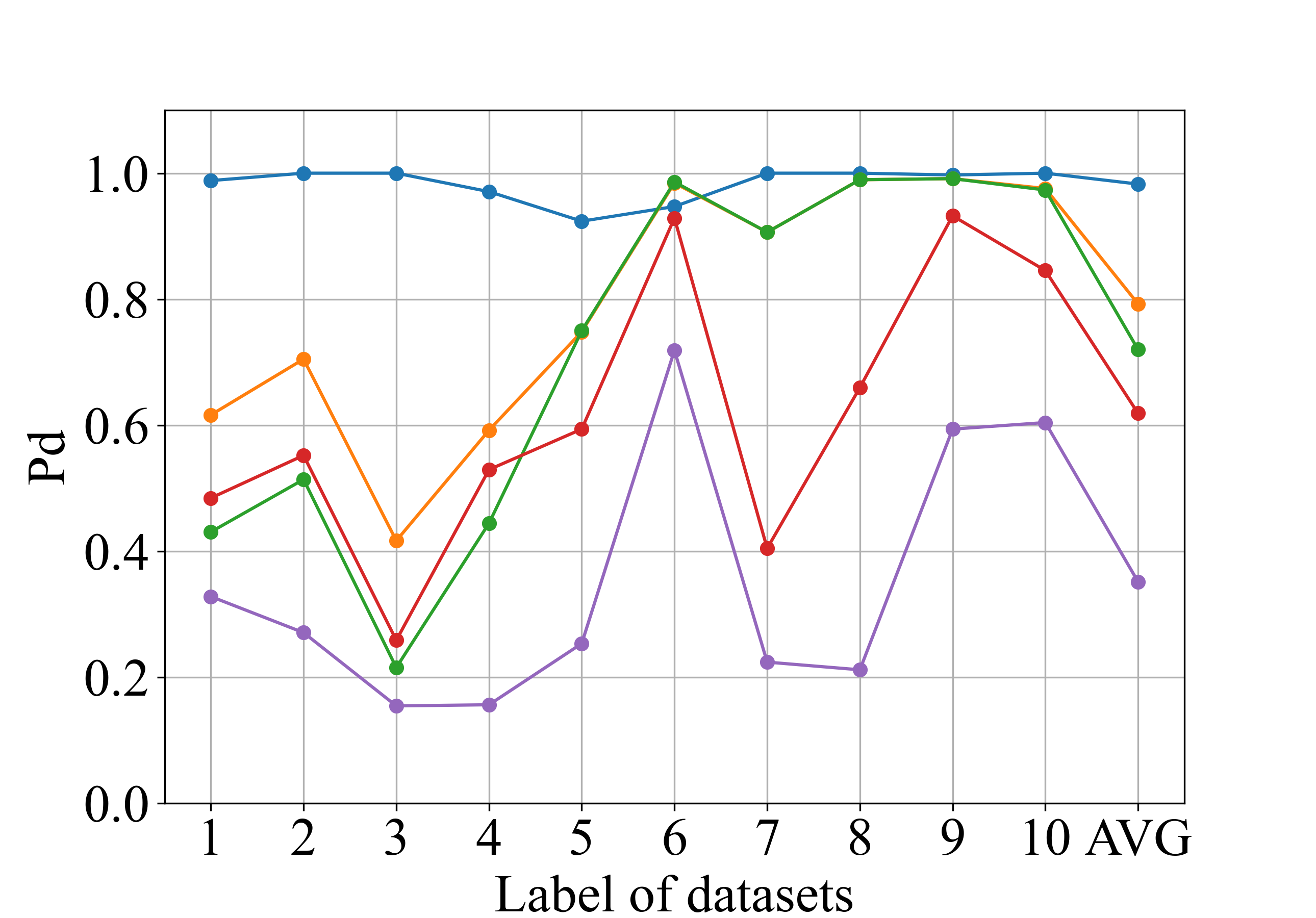}%
  \label{Res_CWT}}
  \hfil
  \subfloat[]{\includegraphics[width=0.5\textwidth]{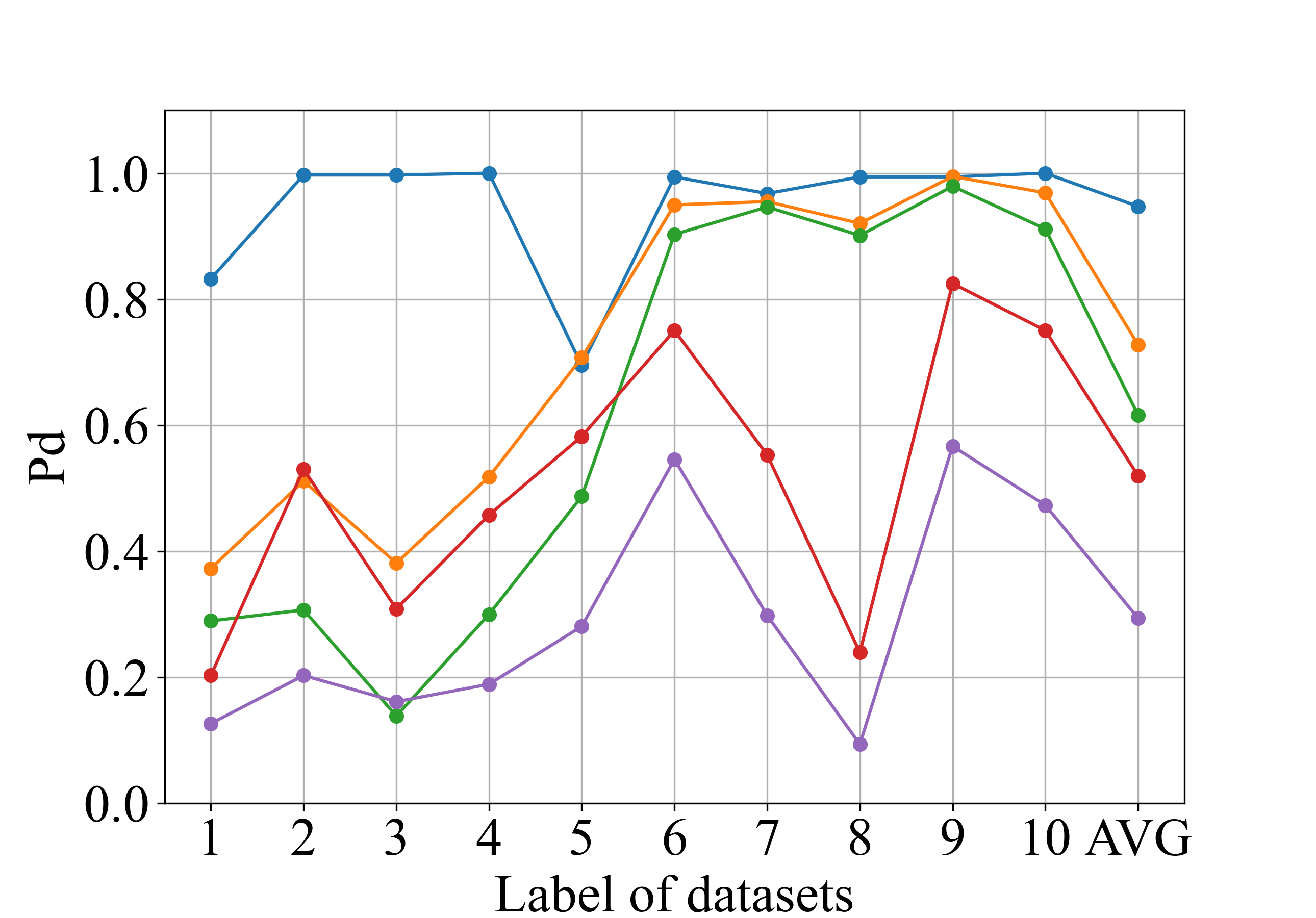}%
  \label{Res_STFT}}
  \hfil
  \includegraphics[width=4in]{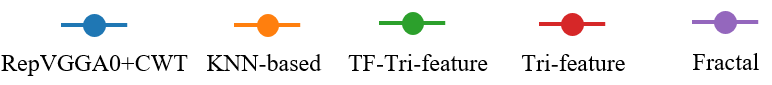}
  \caption{$P_d$ of the RepVGGA0-CWT detector and four existing non-CNN detectors on the ten datasets of IPIX database when target $P_{fa}$ is 0.001 (a) HH. (b) HV. (c) VH. (d) VV.}
  \label{1993_wanhao}
  \end{figure*}

\begin{figure*}[!t]
  \centering
  \subfloat[]{\includegraphics[width=0.33\textwidth]{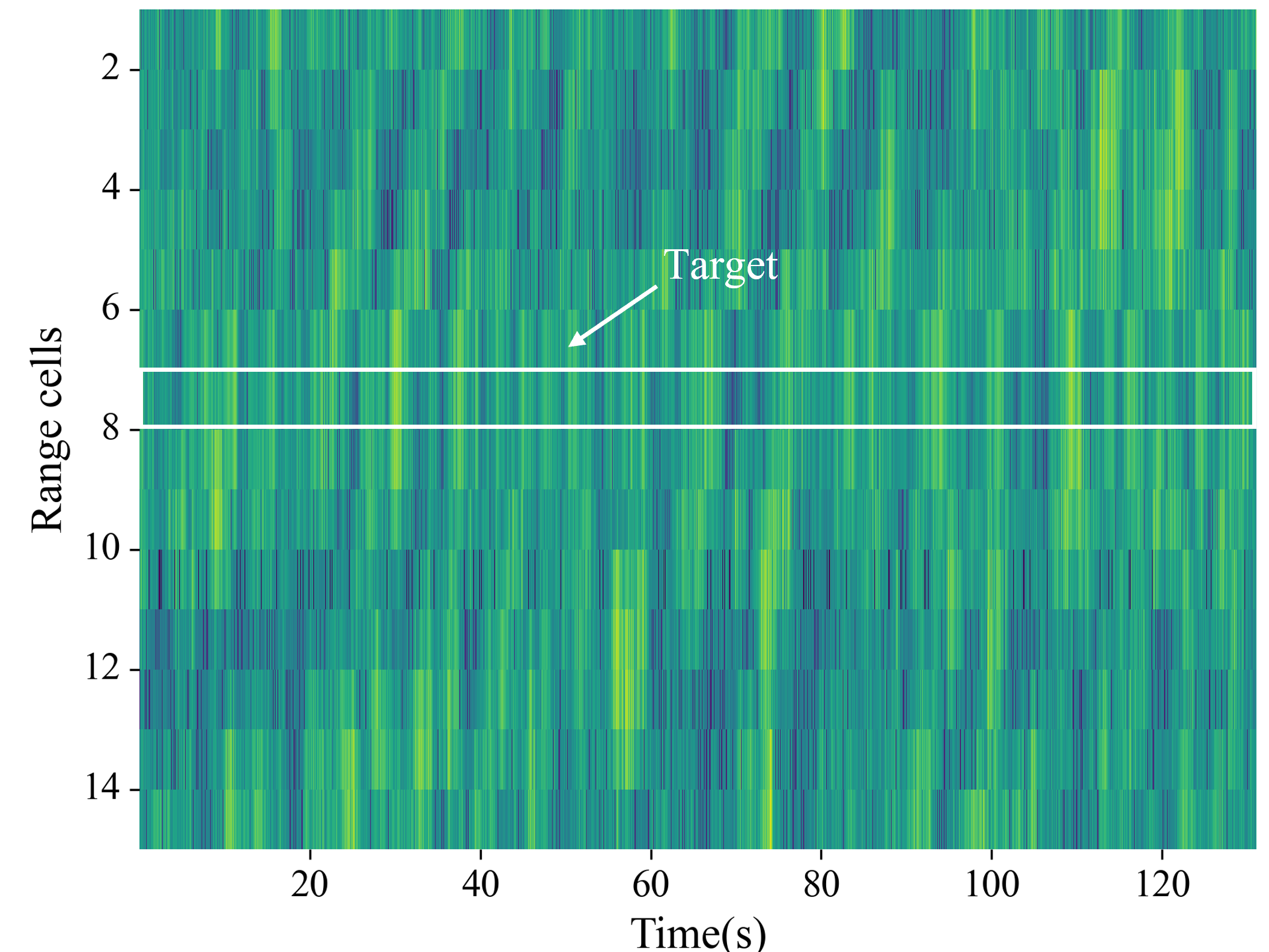}%
  \label{1993_actual_data}}
  \subfloat[]{\includegraphics[width=0.33\textwidth]{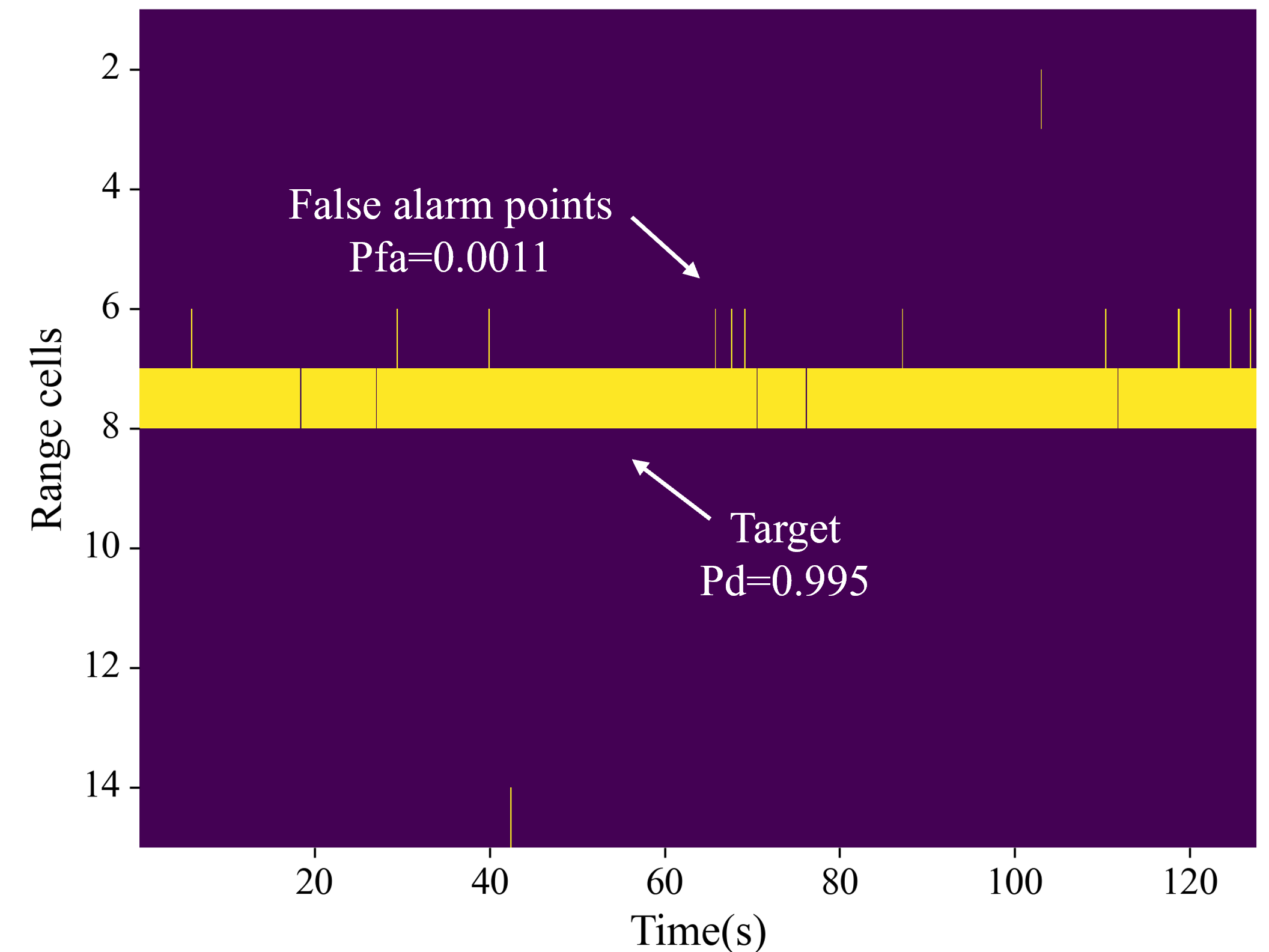}%
  \label{Rep_heatmap1}}
  \subfloat[]{\includegraphics[width=0.33\textwidth]{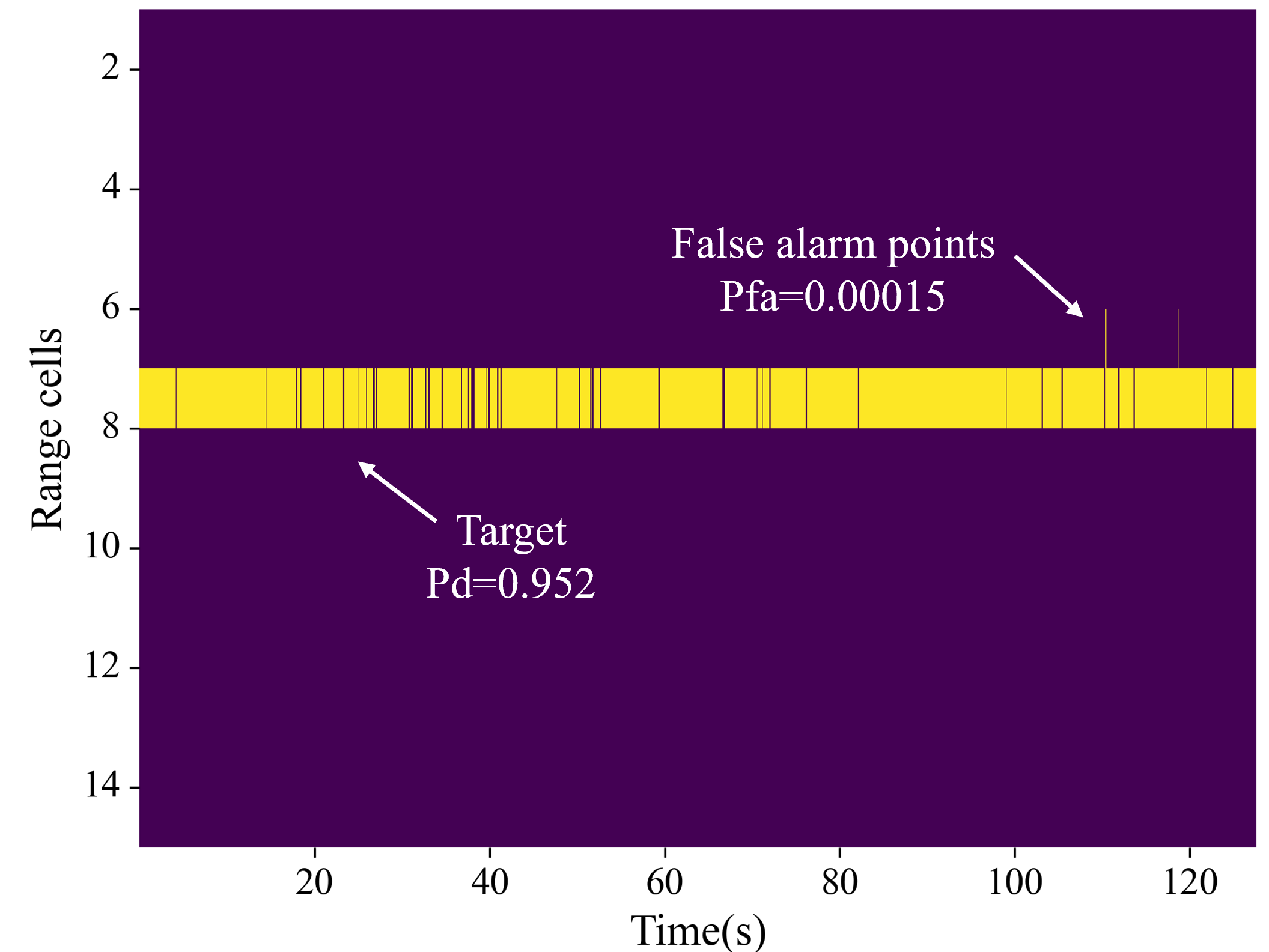}%
  \label{Rep_heatmap2}}
  \hfil
  \subfloat[]{\includegraphics[width=0.33\textwidth]{img/1993raw.png}%
  \label{1993_actual_data}}
  \subfloat[]{\includegraphics[width=0.33\textwidth]{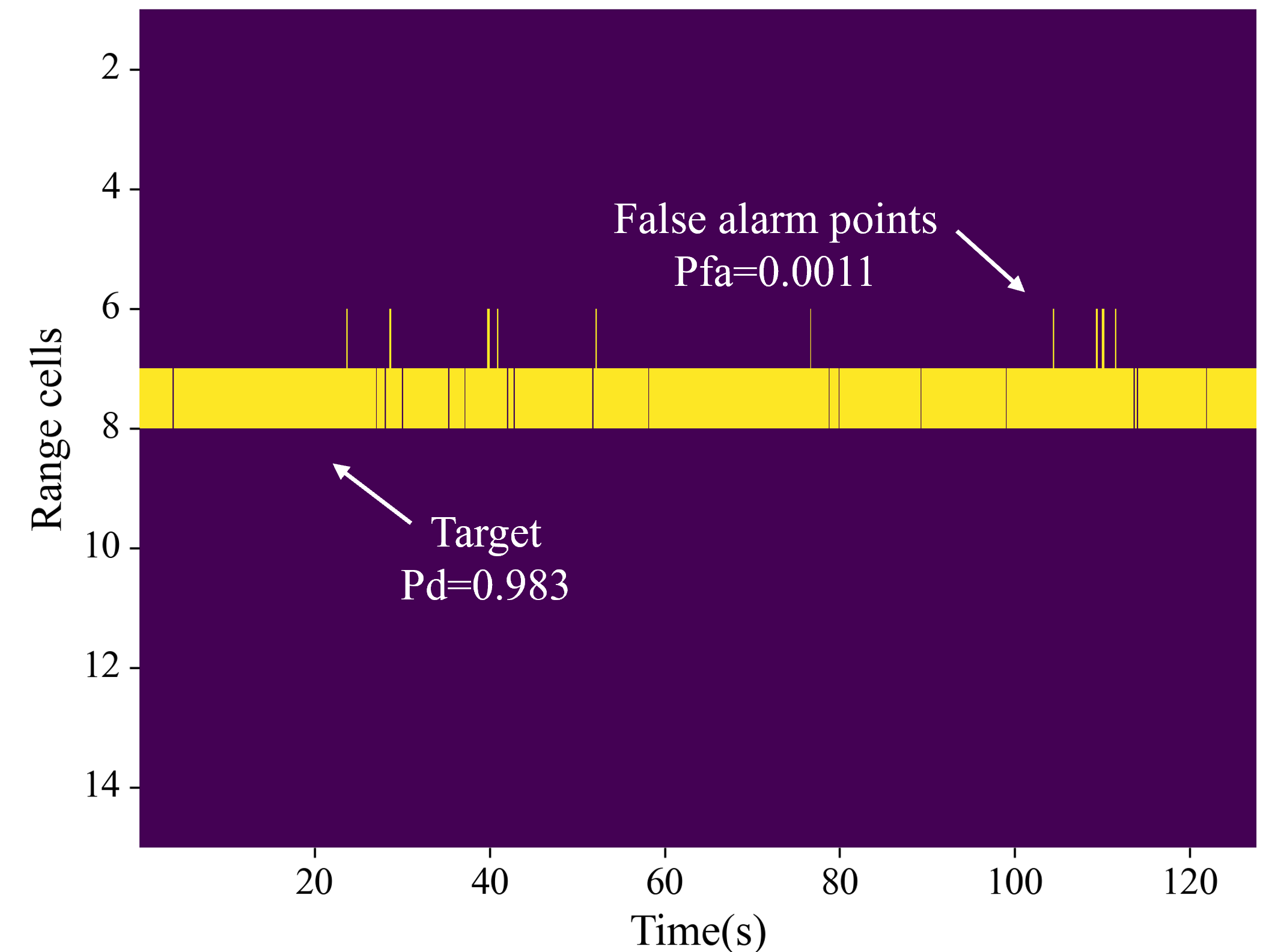}%
  \label{Res_heatmap1}}
  \subfloat[]{\includegraphics[width=0.33\textwidth]{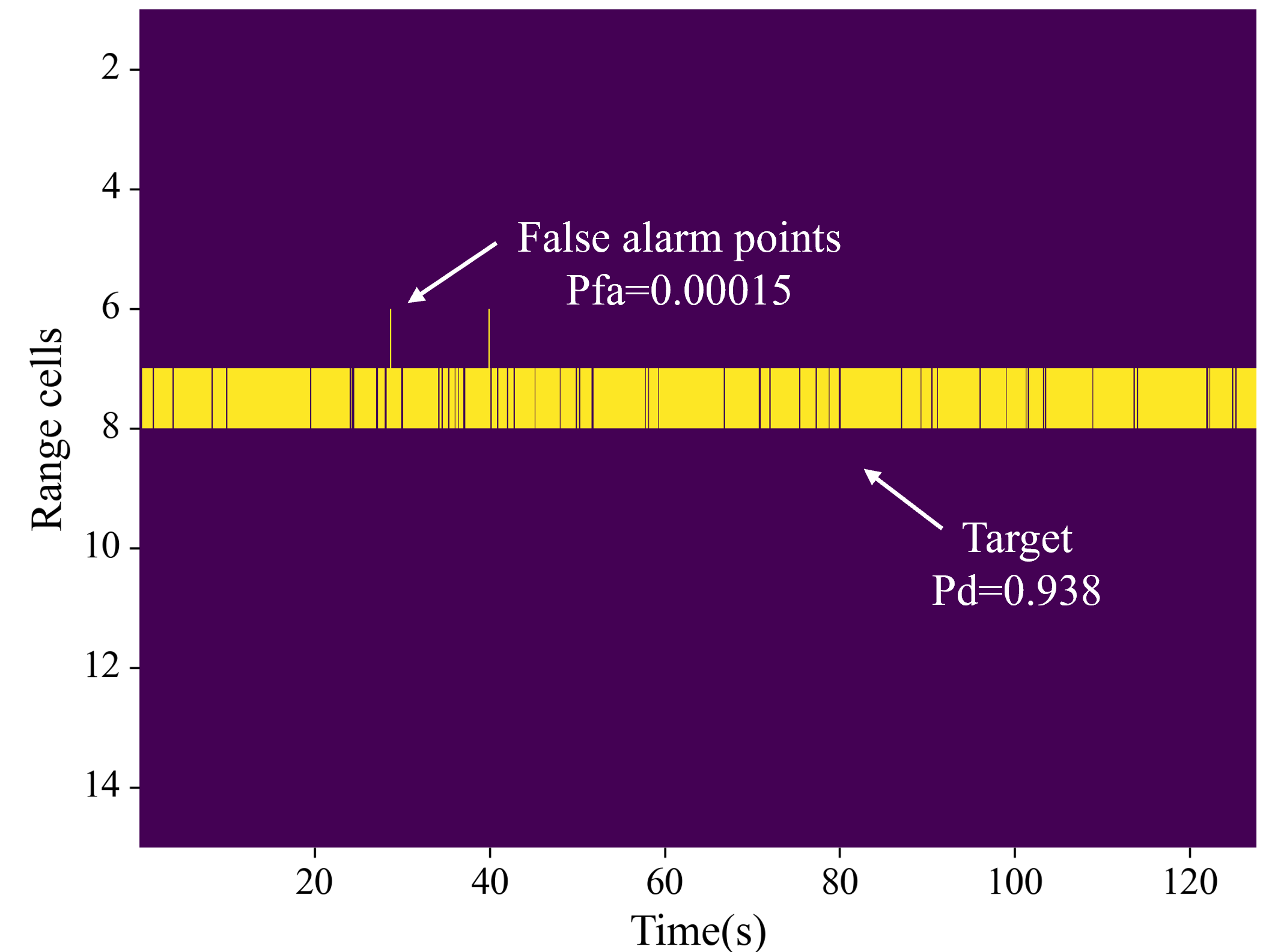}%
  \label{Res_heatmap2}}
  \hfil
  \caption{Detection performance of the RepVGGA0-CWT and the ResNet50-CWT detectors in No.3 IPIX dataset. (a)(d)Actual data. (b)RepVGGA0-CWT, target $P_{fa}$=0.001. (c)RepVGGA0-CWT, target $P_{fa}$=0.0001. (e)ResNet50-CWT, target $P_{fa}$=0.001. (f)ResNet50-CWT, target $P_{fa}$=0.0001}
  \label{heatmap1993}
  \end{figure*}

\begin{figure*}[!t]
  \centering
  \subfloat[]{\includegraphics[width=0.5\textwidth]{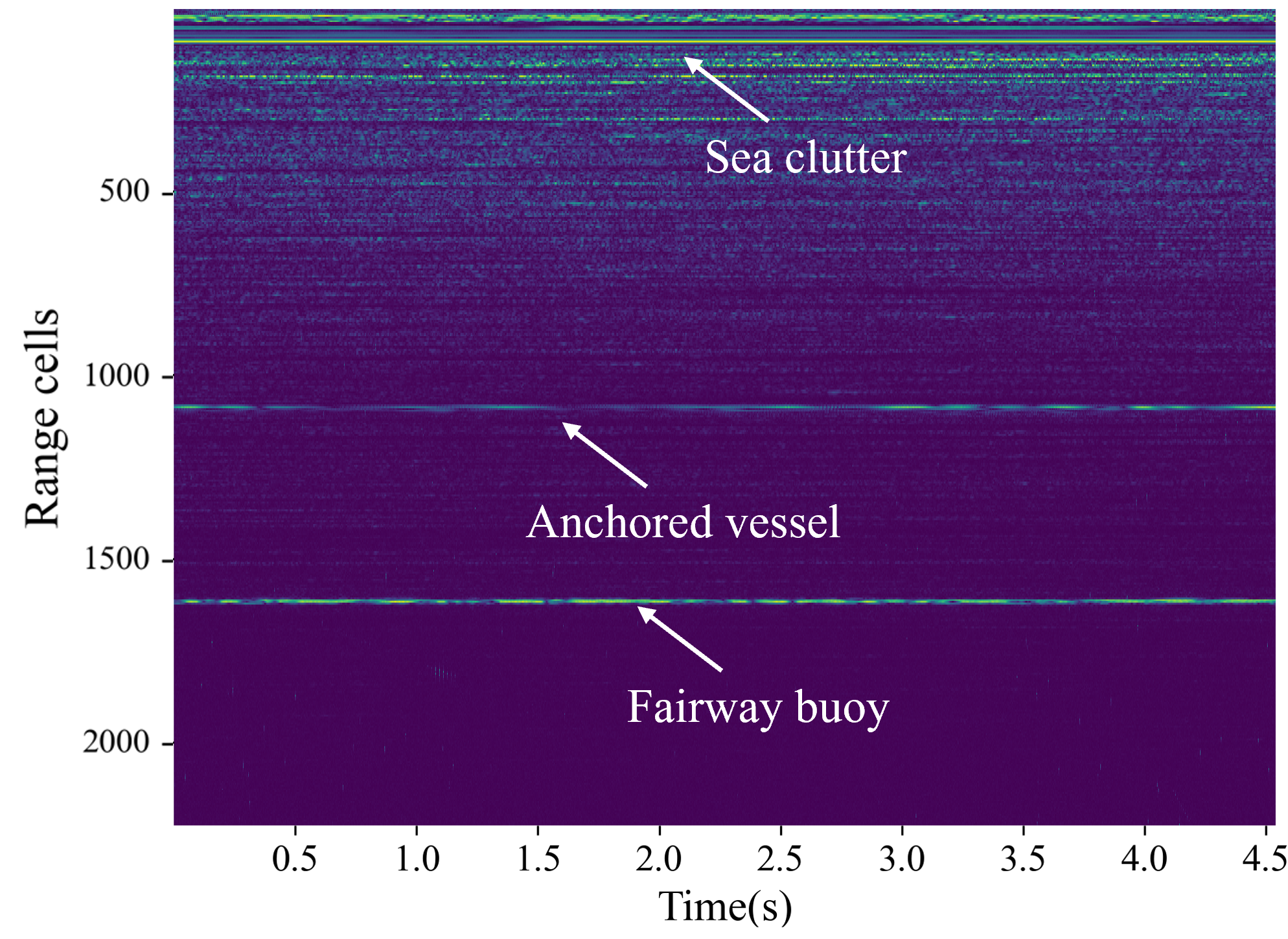}}
  \subfloat[]{\includegraphics[width=0.5\textwidth]{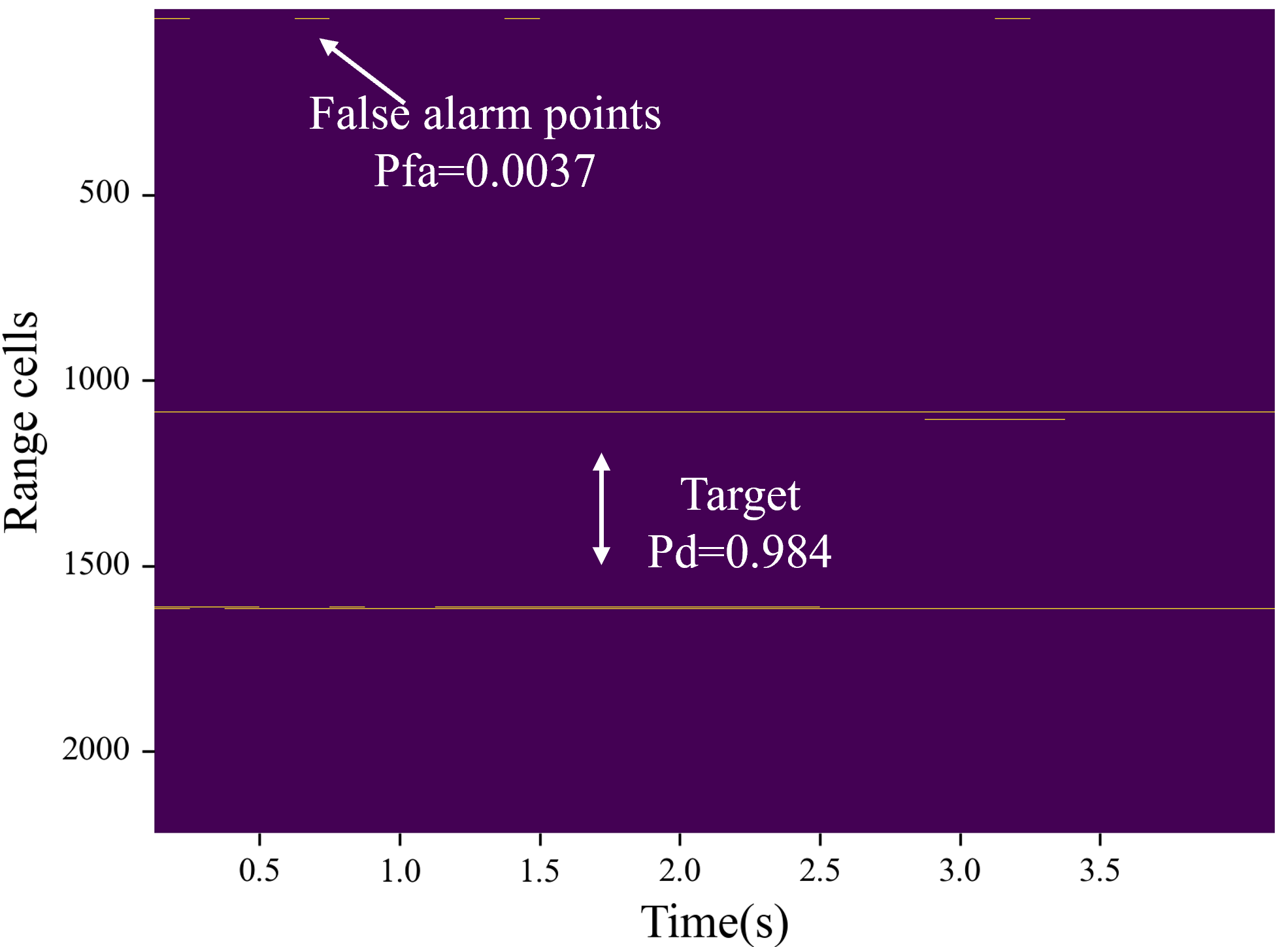}}
  \hfil
  \subfloat[]{\includegraphics[width=0.5\textwidth]{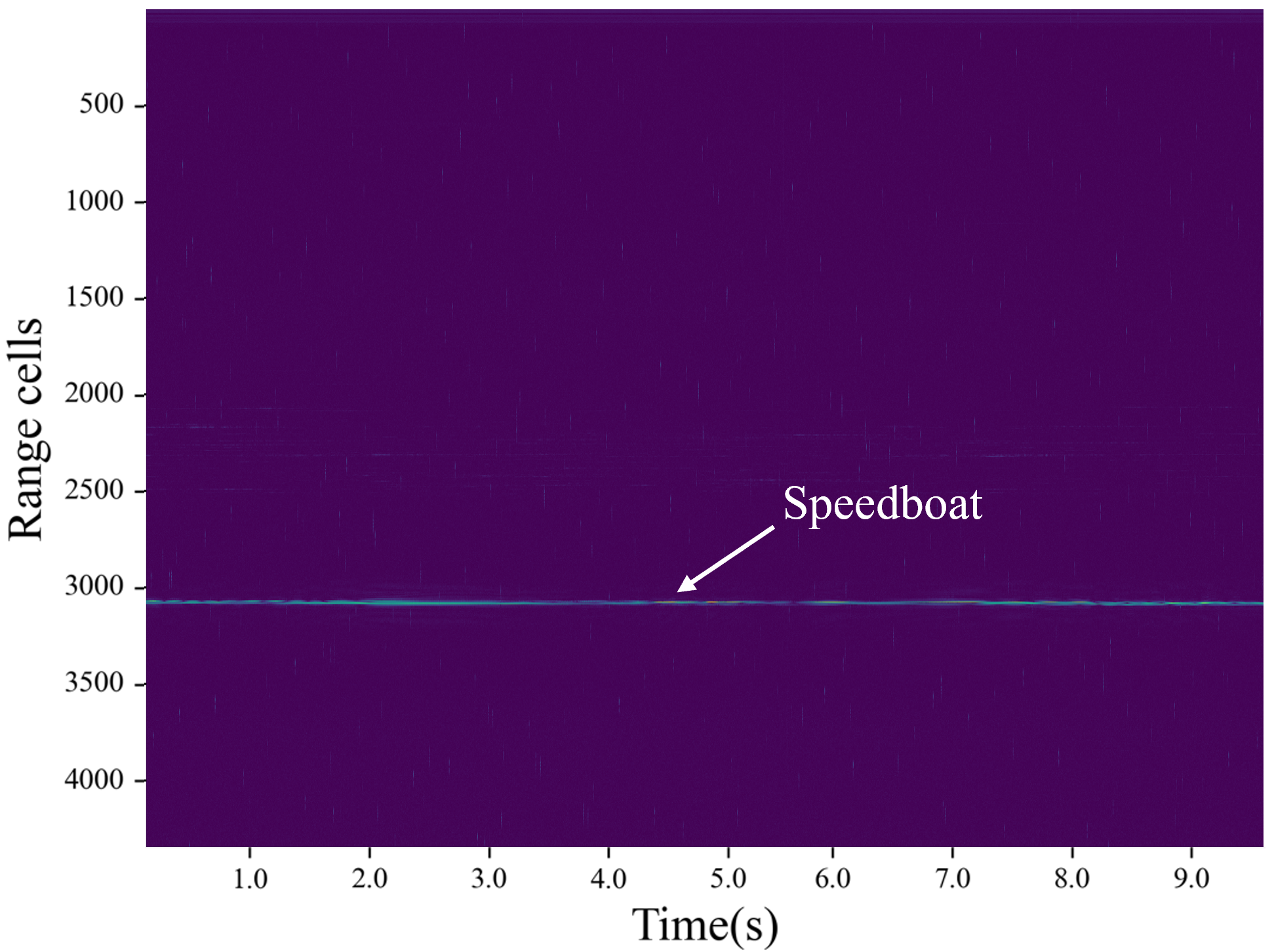}}
  \subfloat[]{\includegraphics[width=0.5\textwidth]{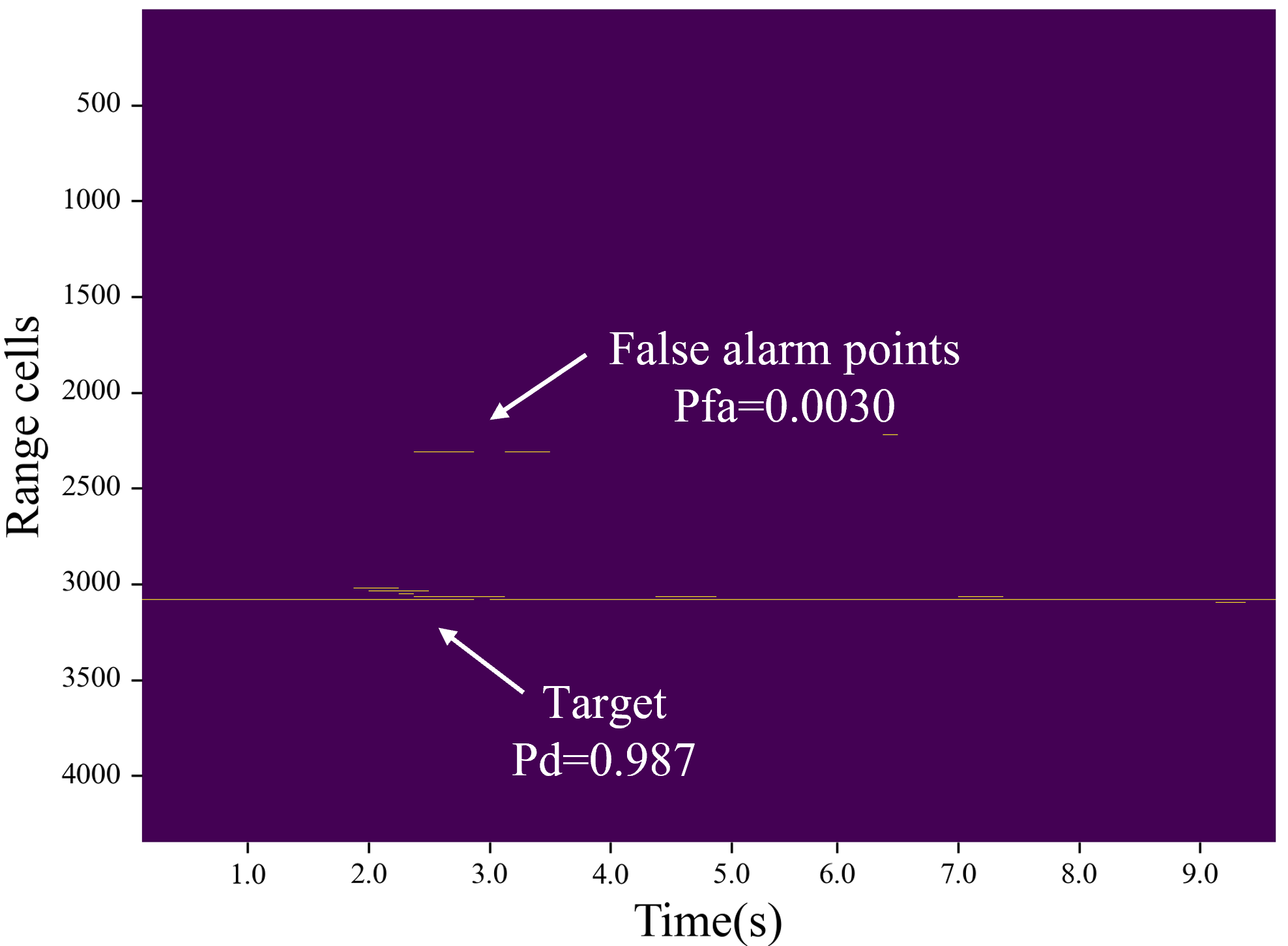}}
  \hfil
  \caption{Detection performance of the RepVGGA0-CWT detector in two 2021 X-band datasets. (a)Actual data of dataset 1. (b)RepVGGA0-CWT detection result of dataset 1. (c)Actual data of dataset 2. (d)RepVGGA0-CWT detection result of dataset 2.}
  \label{heatmap2021}
  \end{figure*}

\section{Results and Comparison}

Before the training of the detector, two significant parameters, determining the construction of training datasets, $N$, the length of the observation window, and $M$, the interval between each observation window, should be decided. A larger $N$ indicates a  longer observation time, which will lead to higher $P_d$ but lower detection speed, while a smaller $N$ can accelerate detection but decrease the accuracy. $M$ determines the number of samples selected from the original IPIX datasets. With fixed $N$, the larger $M$ is, the fewer samples will be obtained. Therefore, we set $N$ to 128, 256, 512, 1024 and set $M$ to 32, 64, 128, 256 and implement a small-scale experiment on No. 4 IPIX dataset to investigate the impact of different $N$ and $M$. The target $P_{fa}$ is set to 0.001 and the result is shown in \textcolor{myblue}{Table} \ref{NMselection}.

\begin{table}[h]
  \centering
  \renewcommand\arraystretch{1.25}
  \caption{detection probability under different $N$ and $M$ when target $P_{fa}$ is set to 0.001}
  \label{NMselection}
\begin{tabular}{ c  c  c  c  c  c }
\toprule

\multirow{2}*{$N$}    & \multirow{2}*{Polarization} & \multicolumn{4}{c}{$M$}             \\ \cmidrule(lr){3-6}
                      &                               & 32     & 64     & 128    & 256     \\ \midrule
\multirow{3}*{128}  & HV,VH                         & 0.4037 & 0.2914 & 0.1930 & 0.2398 \\
                      & HV                            & 0.2491 & 0.1303 & 0.1404 & 0.1520 \\
                      & HH                            & 0.1590 & 0.0556 & 0.0965 & 0.0702 \\ \cmidrule(lr){2-6}
\multirow{3}*{256}  & HV,VH                         & 0.8929 & 0.6701 & 0.5484 & 0.4912 \\
                      & HV                            & 0.9017 & 0.2903 & 0.3402 & 0.2632 \\
                      & HH                            & 0.6456 & 0.1525 & 0.1496 & 0.1287  \\ \cmidrule(lr){2-6}
\multirow{3}*{512}  & HV,VH                         & 1.0000 & 1.0000 & 0.9941 & 0.8304 \\
                      & HV                            & 1.0000 & 0.9985 & 0.9120 & 0.5731  \\
                      & HH                            & 1.0000 & 0.9956 & 0.9384 & 0.2456\\ \cmidrule(lr){2-6}
\multirow{3}*{1024} & HV,VH                         & 1.0000 & 1.0000 & 1.0000 & 1.0000  \\
                      & HV                            & 1.0000 & 1.0000 & 1.0000 & 0.7765  \\
                      & HH                            & 1.0000 & 1.0000 & 1.0000 & 0.9588 \\ \bottomrule
\end{tabular}
\end{table}
As $N$ increases each time, the detector receives doubled radar echo. Consequently, $P_d$ of the detector is improved dramatically and achieves 1.000 when $N$ is set to 512 or larger. In addition, $M$ has less impact on $P_d$ than $N$, especially when $N$ is set to a large value. It can be inferred that in the condition of enough observation time $(N\ge512)$, the performance of detector is able to remain stable even though the number of samples is halved. In order to obtain a detector with both high detection accuracy and fast training speed, the following experiment will be implemented under the condition of $N$ as 512 and $M$ as 128.

From this small-scale experiment, it can also be observed that polarization methods will influence the performance of detector. In order to further investigate the most suitable polarization (or polarization combination) for target detection, another experiment is implemented on each IPIX dataset and target $P_{fa}$ is set to 0.001. There are four detectors in total, including combinations of two feature extraction methods (CWT and STFT) and two neural network structures (RepVGGA0 and ResNet50). \textcolor{myblue}{\textcolor{myblue}{Fig.}} \ref{1993_datasets} illustrates the result of the experiment. Overall, the performance of cross-polarization methods, including HV, VH and their combination, are better than co-polarization methods and $P_d$ of detectors applying CWT are generally higher than $P_d$ of detectors applying STFT. Besides, the RepVGGA0-CWT detector achieves the best performance in average.

In the next experiment, we compared CNN detector with several other non-CNN detectors, and from the performance shown in the \textcolor{myblue}{\textcolor{myblue}{Fig.}} \ref{1993_wanhao}, it is obvious to see that our RepVGG-CWT performs significantly better than other non-CNN detectors on almost all datasets and channels, while controlling the $P_{fa}$.

Especially in cases where clutter and targets are difficult to distinguish (HH data), as mentioned above, thanks to the multi-resolution nature of the detector, it is able to extract more features from the spectrum, resulting in better detection performance.

In \textcolor{myblue}{Table} \ref{polarizationAVG}, exact $P_d$ data of different detectors and polarization methods is shown. An obvious increase in $P_d$ can be observed when the input is extracted by CWT, which produces a 56.9\% improvement in RepVGGA0 and a 49.3\% improvement in ResNet50. On the other hand, the $P_d$ of HV and VH polarization combination is 0.07 higher than cross-polarization methods and 0.2 higher than co-polarization methods. As the result, we consider HV, VH polarization combination as the best polarization method.

\begin{table}[h]
  \tabcolsep=0.1cm
  \centering
  \renewcommand\arraystretch{1.25}
  \caption{detection probability of different detectors and polarization methods when target $P_{fa}$ is set to 0.001}
  \label{polarizationAVG}
\begin{tabular}{ c  c  c  c  c  c  c }
\toprule
Detector      & HV,VH & HV    & VH    & VV    & HH    & AVG    \\ \midrule
\textbf{RepVGGA0-CWT}  & \textbf{0.954} & \textbf{0.940} & \textbf{0.976} & \textbf{0.947} & \textbf{0.928} & \textbf{0.949}  \\
RepVGGA0-STFT & 0.772 & 0.664 & 0.640 & 0.408 & 0.480 & 0.593  \\
ResNet50-CWT  & 0.971 & 0.936 & 0.930 & 0.874 & 0.856 & 0.913  \\
ResNet50-STFT & 0.796 & 0.686 & 0.668 & 0.406 & 0.503 & 0.612  \\
AVG           & 0.873 & 0.806 & 0.804 & 0.659 & 0.692 &      \\ \bottomrule
\end{tabular}
\end{table}

A noteworthy phenomenon is that even though the average $P_d$ of RepVGGA0-STFT detector is lower than the average $P_d$ of ResNet50-STFT detector, the average $P_d$ of RepVGGA0-CWT detector is higher than the average $P_d$ of ResNet50-CWT detector. That may be the evidence of the excellent collaboration between multi-size receptive fields in RepVGGA0 and multi-resolution CWT data. This leads to the superior performance of the RepVGGA0-CWT detector, despite the fact that RepVGGA0 is much smaller than ResNet50. Due to the RepVGG's multi-branch and multi-convolutional kernel network architecture, as compared to the design of ResNet with only residual branches, it is evident that the effect of using multiple convolutional kernel sizes is significantly improved when employing CWT as a multi-resolution feature extraction method.

In particular situation, for example, taking HV, VV or HH polarization as input,the RepVGGA0-CWT detector performs better than the ResNet50-CWT detector. Especially for co-polarization methods, more than 0.07 accuracy improvement is made by RepVGGA0. \textcolor{myblue}{\textcolor{myblue}{Fig.}} \ref{heatmap1993} displays the visualized detection result of No. 3 dataset, where the primary target is located in the 7th distance unit. The results of the target $P_{fa}$ equal to 0.001 detected by the RepVGGA0-CWT and the ResNet50-CWT detectors are shown in \textcolor{myblue}{\textcolor{myblue}{Fig.}} \ref{Rep_heatmap1}, \textcolor{myblue}{\textcolor{myblue}{Fig.}} \ref{Res_heatmap1} and the results of the target $P_{fa}$ equal to 0.0001 is shown in \textcolor{myblue}{\textcolor{myblue}{Fig.}} \ref{Rep_heatmap2}, \textcolor{myblue}{Fig.} \ref{Res_heatmap2}.

Such characteristic also enables the detector to better handle situations where only co-polarization is available. Two recent datasets from a program named “a data-sharing program for sea-detecting radar” are used to test the performance of the detector \cite{XBAND}. Both of them were collected in Yantai, Shandong Province, China and only contain data of HH polarization. In the first dataset (file name: 20210106160919\_01\_staring), an anchored vessel is placed 2.778km (1085th distance unit) away from the radar and a fairway buoy is placed 4.115km (1614th distance unit) from the radar. The second dataset (file name: 20210105160634\_01\_staring) contains a speedboat target 8.15km (3080th distance unit) away from the radar. When the $P_{fa}$ is 0.003, the detection result of RepVGGA0-CWT detector is shown in \textcolor{myblue}{\textcolor{myblue}{Fig.}} \ref{heatmap2021}. In comparison with sequence-feature detector (\cite{BiLSTM}), $P_{d}$ of our detector is improved by 0.256 in the first dataset and 0.075 in the second dataset.

Even though ResNet50-CWT detector performs 0.02 better than RepVGGA0-CWT detector when taking HV, VH polarization combination as input, it should be emphasized that ResNet50 is a comparatively large network, which means that its training and inference will be slow. In the following experiment, the size of trained model and the speed of training and inference of networks are measured. To be more exhaustive, some small networks, including ResNet18 and AlexNet are involved in the experiment. The speed, model size and detection probability of these detectors are shown in \textcolor{myblue}{Table} \ref{NetsOverall}.

\begin{table}[h]
  \tabcolsep=0.1cm
  \centering
  \renewcommand\arraystretch{1.25}
  \caption{speed, model size and detection probability of different detectors when target $P_{fa}$ is set to 0.001, taking HV ,VH polarization combination as input}
  \label{NetsOverall}
\begin{tabular}{ c  c  c  c c }
\toprule

Detector      & $P_d$     & Train FPS & Infer FPS & Model Size  \\ \midrule
AlexNet-CWT   & 0.0704 & 559.3294 & 419.7717 & 87KB \\
AlexNet-STFT  & 0.2362 & 665.3259 & 517.7102 & 87KB \\
ResNet18-CWT  & 0.7540 & 118.8168 & 229.1370 & 85.3MB \\
ResNet18-STFT & 0.7705 & 282.0354 & 514.3297 & 85.3MB \\
RepVGGA0-STFT & 0.7723 & 163.6578 & 344.0007 & 60.0MB \\
ResNet50-STFT & 0.7960 & 79.1786  & 198.5931 & 179MB \\
\textbf{RepVGGA0-CWT}  & \textbf{0.9536} & \textbf{72.3930}  & \textbf{153.7386} & \textbf{60.0MB}\\
ResNet50-CWT  & 0.9707 & 31.1117  & 81.3748  & 179MB \\ \bottomrule
\end{tabular}
\end{table}

In terms of speed comparison, training and inference speed are measured by frame per second (FPS), which indicates how many two-dimensional inputs this detector can process in one second. Detectors with RepVGGA0 are nearly twice as fast as detectors with ResNet50. For training, FPS of the RepVGGA0-CWT detector is 232.69\% as large as FPS of the ResNet50-STFT detector. For inference, FPS of the RepVGGA0-CWT detector is 188.93\% as large as FPS of the ResNet50-STFT detector. In contrast, small networks, like AlexNet and ResNet18, are faster than RepVGGA0, but it is quite difficult for them to obtain a $P_d$ larger or equal to 95\%, which means they cannot be applied in real detection situations. Another noteworthy defect of ResNet is its high memory usage. ResNet50 requires 3 times the space of RepVGGA0 to store its trained model and even the size of trained model of ResNet18 is 1.42 times size of trained model of RepVGGA0. Low memory usage allows RepVGGA0 to be deployed on resource-limited hardware more easily.

\section{Discussion}

The results of experiments show the outstanding ability of RepVGGA0-CWT detector in target detection, which can be attributed to the collaboration between RepVGGA0 and CWT.

Both CWT and STFT can be combined with CNNs for signal processing and feature extraction. However, CWT can provide more time-frequency information, and its frequency resolution is continuous, allowing for more flexible and comprehensive feature extraction through the application of convolution kernels of different sizes at different scales.

To take advantage of such characteristics of CWT, RepVGGA0 is applied. Multiple sizes of convolution kernels in RepVGGA0 increase the network's receptive field and improves its ability to extract features from input images of different scales, allowing the network to learn more diverse features. This feature of RepVGGA0 gives it a significant advantage when processing CWT results.

Therefore, combining CWT and RepVGGA0 better utilizes the time-frequency characteristics of the signal, improving the effectiveness of signal processing and feature extraction, and outperforming the combination of STFT and other CNNs.

However, there are still some deficiencies in this work. The way we generate inputs from CWT results may be quite simplistic. In the situation that $N$ is set to 512, the size of a complete CWT result is 61$\times$512 and the size of an input after even selection is 61$\times$128, which means 75\% information of CWT result is dropped. It is a promising way to increase the detection accuracy if a method that is able to retain more information while reducing the CWT result is proposed.

Even though the RepVGGA0-CWT detector achieves better performance compared with the sequence-feature detector, the training of CNN-based detectors is slower than the training of RNN-based detectors. Considering the high similarity between inputs from different datasets, transfer learning may supersede learning from scratch as a more efficient way to gain an eligible detector.

\section{Conclusion}
This work proposes the RepVGGA0-CWT detector to detect small targets under the background of sea clutter. CWT is applied to extract time-frequency domain features from radar echo and RepVGGA0 is used to detect targets. Compared with typical CNN-based detectors, the proposed RepVGGA0-CWT detector is able to offer high detection accuracy with a very low controllable false alarm rate, high training speed, high inference speed and small memory consumption. 

The RepVGGA0-CWT detector is trained and tested using IPIX datasets and X-band datasets. The $P_d$ of the RepVGGA0-CWT detector achieves 0.949 in average under the condition of target $P_{fa}$ of 0.001, which is 0.036 and 0.318 higher than $P_d$ of the ResNet50-CWT and ResNet50-STFT detector respectively. Moreover, the detector shows an outstanding ability to handle HH and VV polarization radar echo, where a 0.267 improvement in $P_d$ can be observed compared with the sequence-feature detector. 

In this work, we scale down the network input by selecting the points evenly from the result of CWT to accelerate training and inference. It is expected to apply a more advanced method to better reserve the information in time-frequency domain features when the network input is shrunk.

\section*{Declaration of Competing Interest}

The authors declare that they have no known competing financial interests or personal relationships that could have appeared to influence the work reported in this paper.

\section*{Data availability}

The IPIX data used in this work were downloaded from the Dartmouth database (\url{http://soma.mcmaster.ca/ipix/dartmouth/index.html}) and the X-band data used in this work were downloaded from the Journal of Radars (\url{https://radars.ac.cn/web/data/getData?newsColumnId=35452ad7-1048-44c8-9e95-5be1e7d5bdc7}).

\section*{Acknowledgments}

This work was supported in part by the National Natural Science Foundation of China under Grant 61731006, Sichuan Natural Science Foundation under Grant 2023NSFSC0450, and the 111 Project under Grant B17008.The authors would like to thank Ningbo Liu's team for “a data-sharing program for sea-detecting radar” data support.











\bio{}

\endbio


\end{document}